\documentstyle[amsmath,amssymb,jair,epsfig,twoside,11pt,theapa,color,subfig,url]{article}

\ShortHeadings{Finding Still Lifes with Memetic/Exact Hybrid
Algorithms} {Gallardo, Cotta, \& Fern{\'a}ndez} \firstpageno{1}

\begin{document}

\title{Finding Still Lifes with Memetic/Exact Hybrid Algorithms}

\author{\name Jos{\'e} E. Gallardo \email pepeg@lcc.uma.es \\
       \name Carlos Cotta \email ccottap@lcc.uma.es \\
       \name Antonio J. Fern{\'a}ndez \email afdez@lcc.uma.es \\
       \addr Dept. Lenguajes y Ciencias de la Computaci\'on, Universidad de
       M\'alaga,\\
       ETSI Inform\'atica, Campus de Teatinos, 29071 -- M\'alaga,
       Spain}

% For research notes, remove the comment character in the line below.
% \researchnote

\maketitle

%% Para interlineado de 1.5 %%%%%%%%%%%%%%%%%%%%%%%%%%%%%%%%%%%%%%%%
\newlength{\defbaselineskip}
\setlength{\defbaselineskip}{\baselineskip}
\newlength{\defparskip}
\setlength{\defparskip}{1ex}
\newcommand{\setlinespacing}[1]%
{\setlength{\baselineskip}{#1 \defbaselineskip}%
 \setlength{\parskip}{#1 \defparskip}}
%% Fin de para interlineado de 1.5 %%%%%%%%%%%%%%%%%%%%%%%%%%%%%%%%%

% Ecuaciones por casos
\newlength{\svArraycolsep}
\newcommand{\beginCasos}[1]      {{#1}&\,=\,&\left\{\begin{array}{@{}l@{\ \ \ \ }l}}
\newcommand{\beginCasosSinLeft}[1]{{#1}&\,=\,&\left. \begin{array}[t]{@{}l@{\ \ \ \ }l}}
\def\endCasos{\end{array}\right.}

\newenvironment{casosBase}[2]{%
\setlength{\svArraycolsep}{\arraycolsep}%
\setlength{\arraycolsep}{0.0em}%
\begin{eqnarray}%
#1{#2}}%
{\endCasos%
\end{eqnarray}%
\setlength{\arraycolsep}{\svArraycolsep}%
}

\newenvironment{casos}[1]{\begin{casosBase}{\beginCasos}{#1}}{\end{casosBase}\ignorespacesafterend}
\newenvironment{casosSinLeft}[1]{\begin{casosBase}{\beginCasosSinLeft}{#1}}{\end{casosBase}\ignorespacesafterend}

\newcommand{\caso}[2]{{#1},&{#2}\\}

\newcommand{\otroCaso}[1]{%
\endCasos\\%
\beginCasos{#1}%
}
\newcommand{\otroCasoSinLeft}[1]{%
\endCasos\\%
\beginCasosSinLeft{#1}%
}

%%%%%%%%%%%%%%%%%%%%%%%%%%%%%%%%%%%%%%%%%%%%%%%%%%%%%%%%%%%%%%%%%%%%%%%%%%%%%%%%%%%%%%%%%%%%%%%%%
% Entorno para definir algoritmos

\newcounter{AlgCounter}
\newcommand{\LnNumber}{\\[0.1mm] \> \> \small{\theAlgCounter}\,:\' \> \stepcounter{AlgCounter}}
\newcommand{\LnNoNumber}{\\[0.1mm] \> \> \>}
\def\oneTab{\ \ \ \=}
\newenvironment{algorithm}[3]%
{\setcounter{AlgCounter}{1}%
\centering
%\begin{center}%
\rule{0.97\textwidth}{0.75pt}%
%\\\textbf{#1}\\[-0.5\baselineskip]%
%\hrulefill%
%\end{center}%
\vspace*{-0.5\baselineskip}%
\begin{center}%
%\linethickness{1.5pt}
\textbf{#1}\\[-2mm]%
%\hrulefill%
\rule{0.97\textwidth}{0.75pt}%
\end{center}%
\vspace*{-10mm}%
\setlinespacing{0.5}
\begin{tabbing}%
%\ \ \ \ \= \ \ \= \ \ \= \ \ \= \ \ \= \ \ \= \ \ \=\ \ \=\ \ \ \ \ \ \ \ \ \ \ \ \ \ =\kill%
\ \
\oneTab\oneTab\oneTab\oneTab\oneTab\oneTab\oneTab\oneTab\oneTab\oneTab\oneTab\oneTab\oneTab\kill
}%
{\end{tabbing}%
\vspace{-3.5\baselineskip}%
\begin{center}\rule{0.97\textwidth}{0.75pt}\end{center}%
\vspace{-\baselineskip}}

%
%%%%%%%%%%%%%%%%%%%%%%%%%%%%%%%%%%%%%%%%%%%%%%%%%%%%%%%%%%%%%%%%%%%%%%%%%%%%%%%%%%%%%%%%%%%%%%%%%

% Abreviaturas
\def\BB{BnB}
\def\BE{BE}
\def\BS{BS}
\def\BSMABE{\mbox{BS-MA-BE}}
\def\BSMABEMB{\mbox{BS-MA-BE-MB}}
\def\EA{EA}
\def\MA{MA}
\def\MABE{\mbox{MA-BE}}
\def\MABEALLF{\mbox{MA-BE$_{2{\mathrm F}}$}}
\def\MABEONEF{\mbox{MA-BE$_{1{\mathrm F}}$}}
\def\MADLS{\mbox{MA$_{\mathrm{SDLS}}$}}
\def\MATS{\mbox{MA$_{\mathrm{TS}}$}}
\def\MDSLP{MDSLP}
\def\SDLS{SDLS}

\def\embeds{\succ}
\def\split{\gg}
\def\isDef{\triangleq}
\def\occurs{\trianglerighteq}

\def\Adjacents{\mathit{Adjs}}
\def\Adj{\mathit{Adjs'}}
\def\Zeroes{\mathit{zeroes}}
\def\Stable{\mathit{Stable}}

\def\beam{{\cal B}}

\def\dotdot{\,.\,.\,}

\newcommand{\embed}[1]{\widetilde{#1}}
\newcommand{\neighborhood}[3]{{\cal N}({#1,#2,#3})}
\newcommand{\stable}[3]{S(#1,#2,#3)}
\def\funNumNeighbors{\eta}
\newcommand{\numNeighbors}[3]{\funNumNeighbors(#1,#2,#3)}

\newcommand{\porHacer}[1]{\ \\ \hspace*{-0.5cm}{\bf CUIDADO!!!  {\textsc{#1}\\}}}

%
%%%%%%%%%%%%%%%%%%%%%%%%%%%%%%%%%%%%%%%%%%%%%%%%%%%%%%%%%%%%%%%%%%%%%%%%%%%%%%%%%%%%%%%%%%%%%%%%%

\begin{abstract}
The maximum density still life problem (\MDSLP) is a hard constraint
optimization problem based on Conway's game of life. It is a prime
example of weighted constrained optimization problem that has been
recently tackled in the constraint-programming community. Bucket
elimination (\BE) is a complete technique commonly used to solve
this kind of constraint satisfaction problem. When the memory
required to apply BE is too high, a heuristic method based on it
(denominated mini-buckets) can be used to calculate bounds for the
optimal solution. Nevertheless, the curse of dimensionality makes
these techniques unpractical for large size problems. In response to
this situation, we present a memetic algorithm for the \MDSLP\ in
which \BE\ is used as a mechanism for recombining solutions,
providing the best possible child from the parental set.
Subsequently, a multi-level model in which this exact/metaheuristic
hybrid is further hybridized with branch-and-bound techniques and
mini-buckets is studied. Extensive experimental results analyze the
performance of these models and multi-parent recombination. The
resulting algorithm consistently finds optimal patterns for up to
date solved instances in less time than current approaches.
Moreover, it is shown that this proposal provides new best known
solutions for very large instances.
\end{abstract}

\section{Introduction}
\label{stilllife:sect:Introduction}
\label{stilllife:sect:game-of-life}
\label{stilllife:sect:related-work}

The game of life was proposed by John H. Conway in the 60s.
Afterwards, it was divulged by Martin Gardner in his {\em Scientific
American} columns
\cite{gardner:john-conways-new-solitaire-sa70,gardner:cellular-sa71,gardner:wheels-sa83}.
The game is played on an infinite checkerboard in which the only
player places checkers on some of its squares. Each square on the
board is called a cell and has eight neighbors: the eight cells that
share one or two corners with it. A cell is alive if there is a
checker on it, and is dead otherwise. The contents of the board
evolve iteratively, in such a way that the state at time $t$
determines the state at time $t+1$ according to three simple rules
(see Fig.~\ref{stilllife:fig:LifeRules}):
\begin{enumerate}
\item If a cell has exactly two living neighbors, then its
state remains the same in the next iteration. This is called the
{\em life constraint}.
\item If a cell has
exactly three living neighbors, then it is alive in the next
iteration. This is called the {\em birth constraint}.
\item If a cell has fewer than two or more than
three living neighbors, then it is dead in the next iteration. These
are called the {\em death by isolation} and {\em death by
over-crowding} constraints respectively.
\end{enumerate}

\begin{figure}[h!]
    \centering
    \subfloat[][Life]{\label{stilllife:fig:LifeRules:A}\epsfig{file=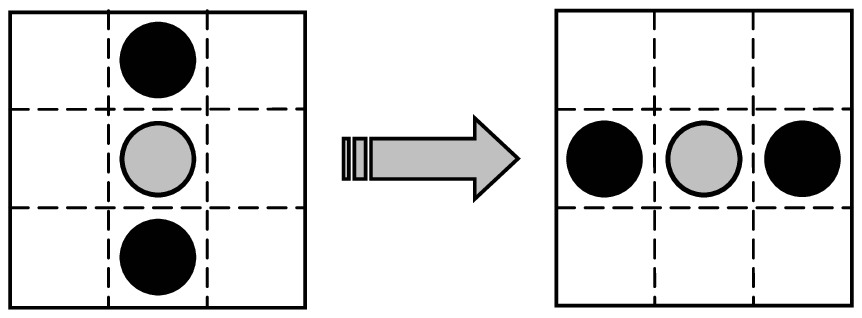,scale=.45}}
    \hspace{20pt}
    \subfloat[][Birth]{\label{stilllife:fig:LifeRules:B}\epsfig{file=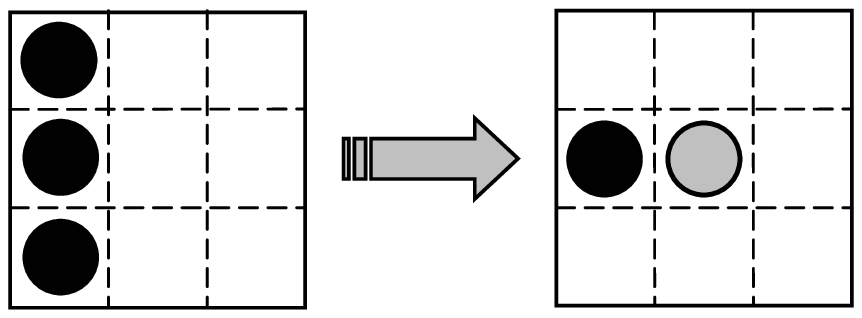,scale=.45}}
    \\
    \vspace{5pt}
    \subfloat[][Death by isolation]{\label{stilllife:fig:LifeRules:C}\epsfig{file=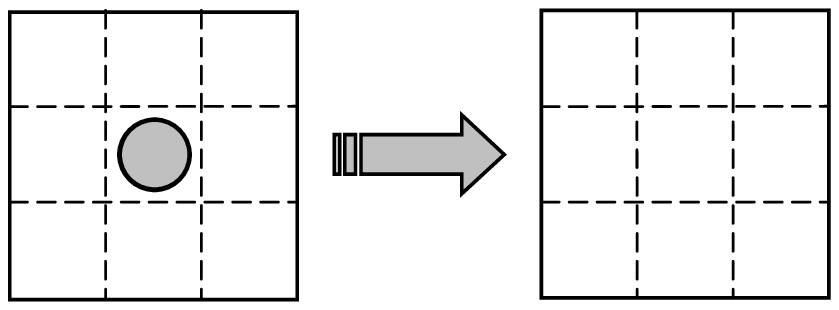,scale=.45}}
    \hspace{20pt}
    \subfloat[][Death by over-crowding]{\label{stilllife:fig:LifeRules:D}\epsfig{file=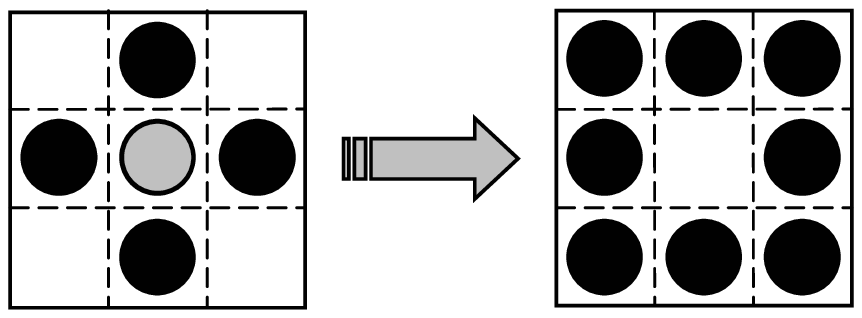,scale=.45}}
    \caption[Rules for the game of life]{ Rules for the Game of Life}
    \label{stilllife:fig:LifeRules}
\end{figure}

As it can be seen, the game of life is defined in terms of simple
rules, but these can nevertheless generate incredibly complicated
patterns and dynamics, and hence, it has attracted the interest of
many scientists.

One challenging constraint optimization problem based on the game of
life is the {\em maximum density still life problem} (\MDSLP). In
order to introduce this problem, let us define a stable pattern
(also called a {\em still life}) as a board configuration that does
not change through  time, and let the {\em density} of a region be
its percentage of living cells. The \MDSLP\ in an $n \times n$ grid
consists of finding a {\em still life} of maximum density.
\citeA{elkies98} has shown that, for infinite boards, the maximum
density is 1/2 (for finite size, no exact formula is known). In this
paper, we are concerned with the \MDSLP\ and finite patterns, that
is, finding maximal $n \times n$ still lifes.

The \MDSLP\ is very hard to solve, and though it has not been
proven to be NP-hard to the best of our knowledge, no
polynomial-time algorithm for it is known. Our interest in this
problem is manifold. Firstly, it must be noted that the patterns
resulting in the game of life are very interesting. For example,
by clever placement of the checkers and adequate interpretation of
the patterns, it is possible to create a Turing-equivalent
computing machine \cite{berlekamp+:winning-math-plays-acpress82}.
From a more applied point of view, it is interesting to consider
that many aspects of discrete dynamical systems have been
developed or illustrated by examples in the game of life
\cite{gardner:cellular-sa71,gardner:wheels-sa83}. In this sense,
finding stable patterns can be regarded as a mathematical
abstraction of a standard issue in discrete systems control.
Finally, the \MDSLP~is a prime example of weighted constrained
optimization problem. As such, it constitutes an excellent test
bed for different optimization techniques. Indeed, the problem has
been included in the CSPLib\footnote{\url{http://www.csplib.org}}
repository.  A dedicated web
page\footnote{\url{http://www.ai.sri.com/~nysmith/life}} maintains
up-to-date results.

The \MDSLP\ has been tackled using different approaches.
\citeA{bosch:three-life-designs-cpaior02} compared different
formulations for the \MDSLP\ using integer programming (IP) and
constraint programming (CP). Their best results were obtained with a
hybrid algorithm mixing the two approaches. They were  able to solve
the cases for $n=14$ and $n=15$ in about 6 and 8 days of CPU time
respectively. \citeA{DBLP:conf/cp/SmithCP02} used a pure constraint
programming approach to undertake the problem and proposed a
formulation of the problem as a constraint satisfaction problem with
0-1 variables and non-binary constraints. Its dual graph translation
into a binary constraint satisfaction problem was also considered.
Surprisingly, it was proven that, although the dual representation
has as many variables as the original one and very large domains,
its performance was much better. However, only instances up to
$n=10$ could be solved. The best results for this problem were
reported by \citeA{DBLP:conf/cp/LarrosaM03} and \citeA{Larrosa05},
showing the usefulness of \emph{bucket elimination} (\BE), an exact
technique based on variable elimination and commonly used for
solving constraint satisfaction problems described in detail in
Section~\ref{stilllife:sect:BE}. Their basic approach could solve
the problem for $n=14$ in about $10^5$ seconds. Further improvements
pushed the solvability boundary forward to $n=20$ in about twice as
much time. Recently,
\citeA{cheng+:global-constraint-still-life-cp05,chengYap06} have
tackled the problem via the use of ad-hoc global {\tt case}
constraints, but their results are comparable to IP/CP hybrids, and
thus lie far from the ones obtained previously by Larrosa et al.

\begin{table}
\caption{Best experimental results reported in
\cite{bosch:three-life-designs-cpaior02} (CP/IP),
\cite{DBLP:conf/cp/LarrosaM03} (BE) and \cite{Larrosa05} (HYB-BE)
for solving the \MDSLP.} \label{stilllife:table:Related:Work}
\centering
\begin{tabular} {rrrrr}\\
    \hline
    $n$   & opt  & CP/IP           &     BE &         HYB-BE \\
    \hline
    12    &   68 &           11536 &   1638 &              1 \\
    13    &   79 &           12050 &  13788 &              2 \\
    14    &   92 & $5 \times 10^5$ & $10^5$ &              2 \\
    15    &  106 & $7 \times 10^5$ &        &             58 \\
    16    &  120 &                 &        &              7 \\
    17    &  137 &                 &        &           1091 \\
    18    &  153 &                 &        &           2029 \\
    19    &  171 &                 &        &          56027 \\
    20    &  190 &                 &        & $2\times 10^5$ \\
    \hline\\
\end{tabular}
\end{table}

Table~\ref{stilllife:table:Related:Work} resumes experimental
results for current approaches used to tackle the \MDSLP. The first
column contains the problem size. The second column shows the
optimal solution as the number of dead cells. Remaining columns
report times in seconds by the hybrid IP/CP algorithm of
\citeA{bosch:three-life-designs-cpaior02}, by the BE approach of
\citeA{DBLP:conf/cp/LarrosaM03} and by the BE/search hybrid of
\citeA{Larrosa05}. Although different computational platforms may
have been used for these experiments, the trends are very clear and
give a pristine indication of the potential of the different
approaches. In this sense, note that all of these techniques applied
to the \MDSLP\ are exact approaches that are inherently limited for
increasing problem sizes, and whose capabilities as anytime
algorithms are unclear. To tackle this problem, we recently proposed
the use of hybrid methods combining exact and metaheuristic
approaches. Particularly, in \cite{GallardoCottaFernandezEVOCOP06}
we considered the hybridization of \BE\ with evolutionary algorithms
(an stochastic population-based search method) endowed with tabu
search (a local search method). The resulting algorithm was a
memetic algorithm (\MA; see Section~\ref{intro:ma}) that used \BE~as
a mechanism for recombining solutions, providing the best possible
child from the parental set. Experimental tests indicated that the
algorithm provided optimal or near-optimal results at an acceptable
computational cost. Afterwards, in \cite{DBLP:conf/ppsn/CottaDFH06}
we studied expanded multi-level models in which our previous hybrid
algorithm was further hybridized with a branch-and-bound derivative,
namely beam search (\BS). Studies about the influence that variable
clustering and multi-parent recombination have on the performance of
the algorithm were also conducted. Results indicated that variable
clustering was detrimental for this problem but also that
multi-parent recombination improves the performance of the
algorithm. To the best of our knowledge, these are the only
heuristic approaches that have been applied to this problem to date.

This paper includes and extends our previous research on this
problem. As new contributions, we have redone all experiments using
an improved implementation of the bucket elimination crossover
operator, described in
Section~\ref{stilllife:sect::SLP:WCSP:modeling}. Additionally, we
present a more extensive experimental analysis of our BS/MA hybrid
described in \cite{DBLP:conf/ppsn/CottaDFH06}, analyzing the
sensitivity of its parameters. We also propose a new hybrid
algorithm that uses the technique of mini-buckets (MB)
\cite{dechter97minibuckets} to further improve the lower bounds of
the partial solutions that are considered in the BS part of the
hybrid algorithm. This new algorithm is obtained from the
hybridization, at different levels, of complete solving techniques
(BE), incomplete deterministic methods (BS and MB) and stochastic
algorithms (MAs). An experimental analysis shows that this new
proposal consistently finds optimal solutions for  MDSLP instances
up to $n=20$ in considerably less time than all the previous
approaches reported in the literature. Finally, in order to test the
scalability of our approach, this novel hybrid algorithm has been
run on very large instances of the MDSLP for which optimal solution
are currently unknown. Results were very successful, as the
algorithm performed at the state-of-the-art, providing solutions
that are equal or better to the best ones reported to date in the
literature.

The paper is self-contained and structured as follows:
Section~\ref{sect:prelimiraries} gives preliminary concepts that
will be used in the rest of paper. Section~\ref{sect:the MDSLP}
defines the MDSLP as a weighted constraint satisfaction problem, and
shows how to solve it using \BE. In Section~\ref{sect:multilevel}, a
MA for the MDSLP that uses BE as a recombination operator is
presented and experimentally evaluated along with a hybrid
multilevel algorithm that integrates the previous MA with
Branch-and-Bound derivatives. Section~\ref{sect:new:hybrid} proposes
and evaluates a novel hybrid algorithm that exploits the technique
of mini-buckets. Finally, Section~\ref{stilllife:sect:conclusions}
presents conclusions and outlines future work.

\section{Preliminaries}
\label{sect:prelimiraries}

In this section, we briefly introduce concepts and techniques that
will be used in the rest of paper. To this end, we first present
beam search, a heuristic tree search algorithm derived from the
branch and bound method. Subsequently, memetic algorithms are
introduced. Finally, weighted constraint satisfaction problems are
defined and the technique of bucket elimination --commonly used to
solve them-- is introduced. For the sake of notational simplicity,
we use in this last subsection the notation of
\cite{DBLP:conf/cp/LarrosaM03,Larrosa05}.

\subsection{Beam Search}
Branch and bound (BB) \cite{LW66} is a general tree search method to
solve combinatorial optimization problems. Tree search methods are
constructive, in the sense that they work on partial solutions. In
this way, tree search methods start with an empty solution that is
incrementally extended by adding components to it. The way that
partial solutions can be extended depends on the constraints imposed
by the problem being solved. The solution construction mechanism
maps the search space to a tree structure, in such a way that a path
from the root of the tree to a leaf node corresponds to the
construction of a solution. In order to efficiently explore this
search tree, BB algorithms maintain an upper bound and estimate
lower bounds for partially constructed solutions. Assuming a
minimization problem, the upper bound corresponds to the cost of the
best solution found so far. During the search process, a lower bound
is computed for any partial solution generated, estimating the cost
of the best solution that can be constructed by extending it. If
this lower bound is greater than the current upper bound, solutions
constructed by extending it will not lead to an improvement, and
thus all nodes descending from it can be pruned from the search
tree. Clearly, the capability of the algorithm for pruning the
search tree depends on the existence of an accurate lower bound,
that should additionally be computationally inexpensive in order to
be practical.

Beam search (BS) \cite{barr81handbookAI} algorithms are incomplete
derivatives of BB algorithms, and are thus heuristic methods.
Essentially, BS works by extending every partial solution from a set
\beam\ (called the {\em beam}) in at most $k_{ext}$ possible ways.
Each new partial solution so generated is stored in a set \beam'.
When all solutions in \beam\ have been processed, the algorithm
constructs a new beam by selecting the best up to $k_{bw}$ (called
the {\em beam width}) solutions from \beam'. Clearly, a way of
estimating the quality of partial solutions, such as a lower bound,
is needed for this.

One interesting peculiarity of BS is that it works by extending in
parallel a set of different partial solutions in several possible
ways. For this reason, BS is a particularly suitable tree search
method to be used in a hybrid collaborative framework, as it can be
used to provide periodically promising partial solutions to a
population based search method such as a MA.
\citeA{GallardoCottaFernandezSMCB} have shown that this kind of
hybrid algorithms can provide excellent results for some
combinatorial optimization problems. We will subsequently present a
hybrid tree search/memetic algorithm for the \MDSLP\ based on this
idea.

\subsection{Memetic Algorithms}
\label{intro:ma}

Evolutionary algorithms (EAs) are population-based metaheuristic
optimization methods inspired by biological evolution
\cite{Bae96,BFM97}. In order to explore the search space, the EA
maintains a set of solutions known as the {\em population} of {\em
individuals} ($\mu$ is used to denote the total number of
individuals in the population). These are initialized usually in a
random way across the search space, although an heuristic may also
be used. After the initialization, three different phases are
iteratively performed until a termination condition is reached: {\em
selection}, {\em reproduction} and {\em replacement}. In the context
of EAs, the objective function assigning values to each solution is
termed a {\em fitness function}, and is used to guide the search by
comparing the goodness of different individuals.

\begin{figure}[t!]
\begin{algorithm}
     {Memetic Algorithm}
     {}
     {}
     \LnNumber \textbf{for} $i :=1$ \textbf{to} $popsize$ \textbf{ do}
     \LnNumber \> $pop[i] :=$ \textsc{Random Board}($n$)
     \LnNumber \> $pop[i] :=$ \textsc{Local Search}($pop[i]$)
     \LnNumber \> \textsc{Evaluate}($pop[i]$)
     \LnNumber  \textbf{end for}
     \LnNumber \textbf{while}  allowed runtime \textbf{not} exceeded  \textbf{do}
     \LnNumber \> \textbf{for} $i := 1$ \textbf{to} \emph{offsize} \textbf{do}
     \LnNumber \> \> \textbf{if} recombination is performed ($p_{X}$) \textbf{then}
     \LnNumber \> \> \> $parent_1 :=$ \textsc{Select}($pop$)
     \LnNumber \> \> \> $parent_2 :=$ \textsc{Select}($pop$)
     \LnNoNumber \> \> \> \dots
     \LnNumber \> \> \> $parent_{arity} :=$ \textsc{Select}($pop$)
     \LnNumber \> \> \> $\mathit{offspring}[i] :=$ \textsc{Recombine}($parent_1$, $parent_2$, \dots, $parent_{arity}$)
     \LnNumber \> \> \textbf{else}
     \LnNumber \> \> \> $\mathit{offspring}[i]:=$ \textsc{Select}($pop$)
     \LnNumber \> \> \textbf{end if}
     \LnNumber \> \> \textbf{if} mutation is performed ($p_m$) \textbf{then}
     \LnNumber \> \> \> $\mathit{offspring}[i]:=$ \textsc{Mutate}($\mathit{offspring}[i]$)
     \LnNumber \> \> \textbf{end if}
     \LnNumber \> \> $\mathit{offspring}[i]:=$ \textsc{Local Search}($\mathit{offspring}[i]$)
     \LnNumber \> \> \textsc{Evaluate}($\mathit{offspring}[i]$)
     \LnNumber \> \textbf{end for}
     \LnNumber \> $pop :=$ \textsc{Replace}($pop$, $\mathit{offspring}$)
     \LnNumber \textbf{end while}
\end{algorithm}
     \caption{Pseudo code of a memetic algorithm (MA). Although different
     variants are possible with respect to this scheme, it broadly captures
     the algorithmic structure typically used in MAs.
     \label{labs:fig:MAcode}}
\end{figure}

Note that EAs are black box optimization procedures in the sense
that no knowledge of the problem (apart from the fitness function)
is used. The need to exploit problem-knowledge has been repeatedly
shown in theory \cite{WM97} and in practice \cite{Dav91} though (see
also \citeR{Cul98_futility}). Different attempts have been made to
answer this need; {Memetic algorithms}
\cite{Moscato_NewMethods_99,Moscato_Cotta_HOM_2002,moscato04memetic,DBLP:journals/tec/KrasnogorS05}
(MAs) are one of the most successful to date \cite{hart05rama}. As
EAs, MAs are also population based metaheuristics. The main
difference is that the components of the population (sometimes
termed {\em agents} in the MAs terminology) are not passive
entities. These agents are active entities that cooperate and
compete in order to find improved solutions.

There are many possible ways to implement MAs. The most common
implementation consists of combining an EA with a procedure to
perform local search on some or all solutions in the population
during the main generation loop (cf.
\citeR{DBLP:journals/tec/KrasnogorS05}). Fig.~\ref{labs:fig:MAcode}
shows the general outline of such a MA. It must be noted however
that the MA paradigm does not simply reduce itself to this
particular scheme and there are diverse places (e.g., population
initialization, genotype to phenotype mapping, evolutionary
operators, etc.) where problem specific knowledge can be
incorporated. In this work, apart form using tabu search
\cite{glover89:tabu:1,glover90:tabu:2}(TS) as a local search
procedure within the MA, we have designed an ``intelligent''
recombination operator that uses an exact technique (bucket
elimination) in order to find the best solution that can be
constructed from a set of parents without introducing implicit
mutation (i.e., exogenous information).

\subsection{Weighted Constraint Satisfaction Problems}

A {\em weighted constraint satisfaction problem} (WCSP)
\cite{schiex95,bistarelli97semiringbased} is a constraint
satisfaction problem (CSP) in which preferences among solutions can
be expressed. Formally, a WCSP can be defined by a tuple $({\cal X},
{\cal D}, {\cal F})$, where ${\cal D}=\{D_1, \cdots, D_n\}$ is a set
of {\em finite domains}, ${\cal X} = \{x_1,\cdots, x_n\}$ is a set
of variables taking values from their finite domains ($D_i$ is the
domain of variable $x_i$) and ${\cal F}$ is a set of {\em cost
functions} (also called {\em soft constraints} or {\em weighted
constraints}) used to declare preferences among possible solutions.
Permitted assignments of variables receive finite costs that express
their degree of preference (the lower the value the better the
preference) and forbidden assignments receive cost $\infty$. Note
that each $f \in {\cal F}$ is defined over a subset of variables,
$var(f) \subseteq {\cal X}$, called its {\em scope}. The objective
function $F$ is defined as the sum of all functions in $\cal{F}$:
\begin{equation}
\label{eq: F}
 F = \sum_{f \in {\cal F}}f
\end{equation}
The assignment of value $v_i \in D_i$ to variable $x_i$ is noted
$x_i = v_i$. A partial assignment of $m$ variables is a tuple $t =
(x_{i_1}=v_1, x_{i_2}=v_2, \cdots, x_{i_m}=v_m$). A complete
assignment of all variables with values in their domains that
satisfies every soft constraint (i.e., with a finite valuation for
$F$) represents a solution to the WCSP. The optimization goal is
to find a solution that minimizes this objective function.

A WCSP\footnote{Observe that the constraints are not weighted in the
sense of having an external weight parameter assigned to them.
Indeed, each of them has the same influence on the overall function
value, as shown in Eq.~(\ref{eq: F}). The reason they are called
``weighted'' is that the output of each function is not binary
(satisfied vs. unsatisfied) but a numerical value when it is
satisfied.} instance is usually depicted by means of its {\em
constraint graph}, which has one node for each variable $x_i \in
{\cal X}$, and one edge connecting any two nodes whose variables
appear in the same scope of some cost function $f \in {\cal F}$.

\subsection{Bucket Elimination}
\label{stilllife:sect:BE}

Bucket elimination (BE) \cite{dechter:bucket-elimination-AI99} is a
generic technique suitable for many automated reasoning and
optimization problems and, in particular, for WCSP solving. The
functioning of BE is based upon the following two operators over
functions \cite{Larrosa05}:

\begin{itemize}
\item {\em the sum of two functions} $f$ and $g$, denoted $(f + g)$, is a new
function with scope $var(f) \cup var(g)$ which returns for each
tuple the sum of costs of $f$ and $g$,
\begin{equation}
(f + g)(t) = f(t) + g(t).
\end{equation}
\item The elimination of variable $x_i$ from $f$, denoted $f \Downarrow x_i$, is a new
function with scope $var(f) - \{x_i\}$ which returns for each
tuple $t$ the minimum cost extension of $t$ to $x_i$,
\begin{equation}
(f \Downarrow x_i)(t) = min_{v \in D_i}  \{f(t \cdot (x_i = v))\},
\end{equation}
where $t \cdot (x_i = v)$ means the extension of the assignment $t$
with the assignment of value $v$ to variable $x_i$. Observe that
when $f$ is a unary function (i.e., it has arity one), a constant is
obtained upon elimination of the only variable in its scope.
\end{itemize}

\begin{figure}[h!]
\begin{algorithm}
{Bucket Elimination for a WCSP $({\cal X},{\cal D},F)$}{}{}
     \LnNoNumber {\bf function} BE(${\cal X},{\cal D},{\cal F}$)
     \LnNumber {\bf for} $i := n$ {\bf downto} $1$ {\bf do}
     \LnNumber \> $B_i := \{f \in {\cal F} \mid x_i \in var(f)\}$
     \LnNumber \> $g_i := (\sum_{f \in B_i} f) \Downarrow x_i$
     \LnNumber \> ${\cal F} := ({\cal F} \bigcup \{g_i\}) - B_i$
     \LnNumber {\bf end for}
     \LnNumber $t := \emptyset$
     \LnNumber {\bf for} $i := 1$ {\bf to} $n$ {\bf do}
     \LnNumber \> $v := argmin_{a \in D_i}\{(\sum_{f \in B_i} f)(t \cdot (x_i = a))\}$
     \LnNumber \> $t := t \cdot (x_i = v)$
     \LnNumber {\bf end for}
     \LnNumber {\bf return}$({\cal F},t)$
     \LnNoNumber {\bf end function}
\end{algorithm}
\caption[General template, adapted from
\citeA{DBLP:conf/cp/LarrosaM03}, of bucket elimination for a WCSP
$({\cal X},{\cal D},F)$.]{The general template, adapted from
\citeA{DBLP:conf/cp/LarrosaM03}, of bucket elimination for a WCSP
$({\cal X},{\cal D},F)$.} \label{stilllife:fig:be:wcsp}
\end{figure}

Without loss of generality, let us assume a lexicographical ordering
for the variables in ${\cal X}$, i.e., $o = (x_1, x_2, \cdots,
x_n)$. Fig.~\ref{stilllife:fig:be:wcsp} shows a pseudo-code of the
BE algorithm for solving a WCSP instance, that returns the optimal
cost in ${\cal F}$ and one optimal assignment in $t$. Observe that,
in a first phase, BE eliminates one variable $x_i \in {\cal X}$ in
each iteration of the loop comprising lines 1-5. This is done by
computing firstly the bucket $B_i$ of variable $x_i$ as the set of
all cost functions in ${\cal F}$ having $x_i$ in their scope. Then,
a new function $g_i$ is defined as the sum of all these functions in
$B_i$ in which variable $x_i$ has been eliminated. Finally, ${\cal
F}$ is updated by removing the functions involving $x_i$ (i.e.,
those in $B_i$) and adding the new function that does not contain
$x_i$. The consequence is that $x_i$ does not exist in ${\cal F}$
but the value of the optimal cost is preserved. The elimination of
$x_1$ produces an empty scope function (i.e., a constant) which is
the optimal cost of the problem. Then, in lines 6-10, BE generates
an optimal assignment of variables by considering these in the order
imposed by $o$: this is done by starting from an empty assignment
$t$ and assigning to $x_i$ the best value regarding the extension of
$t$ with respect to the sum of functions in $B_i$
($argmin_a\{f(a)\}$ denotes the value of $a$ producing minimum
$f(a)$).

Note that BE has exponential space complexity because, in general,
the result of summing functions or eliminating variables cannot be
expressed intensionally by algebraic expressions and, as a
consequence, intermediate results have to be collected extensionally
in tables. To be precise, the complexity of BE depends on the
problem structure (as captured by its constraint graph $G$) and the
ordering $o$. According to \citeA{DBLP:conf/cp/LarrosaM03}, the
complexity of BE along ordering $o$ is time $\Theta(Q \times n
\times d^{w^{*}(o)+1})$ and space $\Theta(n \times d^{w^{*}(o)})$,
where $d$ is the largest domain size, $Q$ is the cost of evaluating
cost functions (usually assumed $\Theta(1)$), and $w^*(o)$ is the
{\em induced width} of the graph along ordering $o$, which describes
the largest clique created in the graph by bucket elimination, and
which corresponds to the largest scope of a function recorded by the
algorithm. Although finding the optimal ordering is NP-hard
\cite{Arnborg85}, heuristics and approximation algorithms have been
developed for this task (check
\citeR{dechter:bucket-elimination-AI99} for details).

\section{The Maximum Density Still Life Problem}
\label{sect:the MDSLP} According to the definition of \MDSLP\ and
the three rules of the game, it is easy to see that each cell in a
still life must satisfy the following conditions:
\begin{itemize}
\item If the cell is alive, it must have two or three neighbors.
\item If the cell is dead, it will have either more than three or less than three neighbors.
\end{itemize}
Note that finite still lifes are not allowed to produce new living
cells outside the grid, and hence stability conditions must hold in
the cells surrounding the $n \times n$ square, that are assumed to
be dead. This can equally be achieved by requiring further that:
\begin{itemize}
\item if the cell is at the boundary of the $n \times n$ square, it must not be part of a sequence
of three consecutive living cells in the direction of the boundary.
\end{itemize}

Fig.~\ref{stilllife:fig:MDSLsExamples} shows some maximum density
still lifes for small values of $n$.

\begin{figure}[h!]
   \centering \epsfig{file=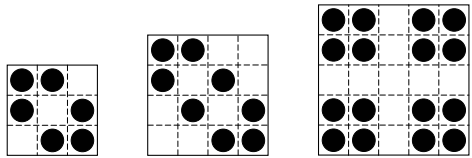, scale=.40}
   \caption{Maximum density still lifes for $n \in \{3,4,5\}$.}
   \label{stilllife:fig:MDSLsExamples}
\end{figure}

The constraints and objectives of the \MDSLP\ are formalized in the
following subsections in which we follow a similar notation to the
one used in \cite{DBLP:conf/cp/LarrosaM03,Larrosa05}.

\subsection{Problem Formulation}
\label{subsect:Problem Formulation}

To state the problem formally, let $r$ be an $n \times n$ binary
matrix, such that $r_{ij} \in\{0,1\}, 1 \leqslant i,j \leqslant n$
($r_{ij}=0$ if cell $(i,j)$ is dead, and 1 otherwise). In addition,
let $\neighborhood{r}{i}{j}$ be the set comprising the neighborhood
of cell $r_{ij}$:
\setlength\svArraycolsep{\arraycolsep}%
\setlength{\arraycolsep}{0.0em}%
\begin{eqnarray}
\neighborhood{r}{i}{j} = \{\,r_{(i+x)(j+y)}\, |\, x,y \in \{-1,0,1\}
\wedge x^2+y^2 \neq 0  \mbox{} \wedge\ 1 \leqslant (i+x), (j+y)
\leqslant \|r\|\, \}
\end{eqnarray}%
\setlength{\arraycolsep}{\svArraycolsep}%
where $\|r\|$ denotes the number of rows (or columns) of square
matrix $r$, and let the number of living neighbors for cell $r_{ij}$
be noted as $\numNeighbors{r}{i}{j}$:
\begin{equation}
\label{stilllife:eq:numNeighbors} \numNeighbors{r}{i}{j} = \sum_{c
\in \neighborhood{r}{i}{j}} c
\end{equation}

According to the rules of the game, let us also define the following
predicate that checks whether cell $r_{ij}$ is stable:
\begin{casos}
{\stable{r}{i}{j}}
 \caso{2 \leqslant \numNeighbors{r}{i}{j} \leqslant 3}{r_{ij} = 1}
 \caso{\numNeighbors{r}{i}{j} \neq 3}{r_{ij} = 0.}
\end{casos}

In order to check boundary conditions, we will further denote by
$\embed{r}$  the $(n+2) \times (n+2)$  matrix obtained by embedding
$r$ in a frame of dead cells:
\begin{casos}
{\embed{r}_{ij}}\label{stilllife:eq:embed}
 \caso{r_{(i-1) (j-1)}}{2 \leqslant i,j \leqslant n+1}
 \caso{0}{\rm{otherwise}.}
\end{casos}

The maximum density still life problem for an $n \times n$ board,
\MDSLP$(n)$, can now be stated as finding an $n \times n$ binary
matrix $r$, such that
\begin{eqnarray}
\sum_{1 \leqslant i,j \leqslant n}(1-r_{ij}) {\rm\ is\ minimal,}
\end{eqnarray}
subject to
\begin{eqnarray}
\bigwedge_{1 \leqslant i,j \leqslant n+2} \stable{\embed{r}}{i}{j}.
\label{stilllife:eq:MDSLPConstraints}
\end{eqnarray}

\subsection{The \MDSLP\ as a Weighted Constraint Satisfaction Problem}
\label{stilllife:sect::SLP:WCSP:modeling}

As shown by \citeA{DBLP:conf/cp/LarrosaM03} and \citeA{Larrosa05},
the \MDSLP\ fits nicely within the framework of WCSPs. To this end,
an $n \times n$ board configuration can be represented by an
$n$-dimensional vector $(r_1,r_2,\dots,r_n)$. Each vector component
encodes (as a binary string) a row, so that the $j$-th bit of row
$r_i$ (noted $r_{ij}$) signifies the state of the $j$-th cell of the
$i$-th row (a value of 1 represents a live cell and a value of 0 a
dead cell).

Two functions over rows will be useful to describe the constraints
that must be satisfied by a valid configuration. The first one,
\begin{equation}
\Zeroes(a) = \sum_{1 \leqslant i \leqslant n} (1 - a_i),
\end{equation}
returns the number of dead cells in a row (i.e., the number of
zeroes in binary string $a$). The second one,
\begin{casosSinLeft}
{\Adjacents(a)}{\Adj(a,1,0)} \otroCaso{\Adj(a,i,l)\nonumber}
 \caso{l}{i > n}
 \caso{\Adj(a,i+1,l+1)}{a_i = 1}
 \caso{\max(l,\Adj(a,i+1,0))}{a_i = 0,}
\end{casosSinLeft}
computes the maximum number of adjacent living cells in row $a$. We
also introduce a ternary predicate, $\Stable(r_{i-1},r,r_{i+1})$,
that takes three consecutive rows in a board configuration  and is
satisfied if, and only if, all cells in the central row are stable
(i.e., all cells in $r$ will remain unchanged in the next
iteration):
\begin{casosSinLeft}
{\Stable(a,b,c)}{\bigwedge_{1 \leqslant i \leqslant n} S(a,b,c,i)}
\otroCaso{S(a,b,c,i)\nonumber}
 \caso{2 \leqslant \eta(a,b,c,i) \leqslant 3}{b_i =1}
 \caso{\eta(a,b,c,i) \neq 3}{b_i = 0}
\otroCasoSinLeft{\eta(a,b,c,i)\nonumber}{\begin{array}[t]{l}
                                         \sum_{\max(1,i-1)\leqslant j \leqslant \min(n,i+1)}(a_j+b_j+c_j) - b_i,
                                         \end{array}}
\end{casosSinLeft}
where $\eta(a,b,c,i)$ is the number of living neighbors of cell
$b_i$, assuming $a$ and $c$ are the rows above and below row $b$.

The \MDSLP\ can now be formulated as a WCSP using $n$ cost functions
$f_i$, $i \in \{1\dotdot n\}$. Accordingly, $f_n$ is binary with
scope the last two rows of the board ($var(f_n) = \{r_{n-1},r_n\}$)
and is defined as:
\begin{casos}
{f_n(a,b)} \caso{\infty}{\neg \Stable(a,b,0)}
\caso{\infty}{\Adjacents(b) > 2} \caso{\Zeroes(b)}{\rm{otherwise.}}
\end{casos}
The first line checks that all cells in row $r_n$ are stable,
whereas the second one checks that no new cells are produced below
the $n \times n$ board. Note that any pair of rows representing an
unstable configuration is assigned a cost of $\infty$, whereas a
stable one is assigned its number of dead cells (to be minimized).

For  $i \in \{2 \dotdot n-1\}$, corresponding $f_i$ cost functions
are ternary with scope $var(f_i) = \{r_{i-1},r_i,r_{i+1}\}$ and are
defined as:
\begin{casos}
{f_i(a,b,c)} \caso{\infty}{\neg \Stable(a,b,c)} \caso{\infty}{a_1 =
b_1 = c_1 = 1} \caso{\infty}{ a_n = b_n = c_n = 1}
\caso{\Zeroes(b)}{\rm{otherwise.}}
\end{casos}
In this case, boundary conditions are checked to the left and right
of the board. As regards cost function $f_1$, it is binary with
scope the first two rows of the board ($var(f_1) = \{r_1,r_2\}$) and
is specified similarly to $f_n$:
\begin{casos}{f_1(b,c)}
\caso{\infty}{\neg \Stable(0,b,c)} \caso{\infty}{\Adjacents(b) > 2}
\caso{\Zeroes(b)}{\rm{otherwise.}}
\end{casos}

\subsection{Solving the \MDSLP\ with BE} \label{stilllife:sect::SLP:WCSP}
According to the formulation of the \MDSLP\ as a WSCP introduced in
Section~\ref{stilllife:sect::SLP:WCSP:modeling}, the corresponding
constraint graph has a sequential structure, in which an arbitrary
row is linked to the two rows above and below it. Due to this simple
structure, it is easy to find an optimal elimination order for BE,
and variables can be eliminated starting with the last one and
proceeding in decreasing order.
Fig.~\ref{stilllife:fig:be:algorithm:mdslp} shows the resulting
algorithm. Function \BE~takes two parameters: $n$, the size of the
instance to be solved, and ${\cal D}$, the domain for each variable
(row) in the solution. If domain ${\cal D}$ is set to $\{0 \dotdot
2^n-1\}$ (i.e., a set containing all possible rows) the function
implements an exact method that returns the optimal solution for the
problem instance (as the number of dead cells) and a vector
corresponding to the rows of that solution. The algorithm starts by
eliminating the last variable $r_n$, whose bucket is $B_n =
\{f_n,f_{n-1}\}$, the only cost functions containing $r_n$ in their
scopes. In lines 1-3, $B_n$ is used to compute a new cost function,
$g_n(a,b)$, with scope $\{r_{n-2},r_{n-1}\}$, that represents the
cost of the best extension of $(r_{n-2}=a, r_{n-1}=b)$ to the
removed variable $r_n$. At this point, the bucket of the next
variable, $r_{n-1}$, is $B_{n-1}=\{g_n,f_{n-2}\}$, that can be used
to compute a new cost function, $g_{n-1}(a,b)$ with scope
$\{r_{n-3},r_{n-2}\}$ representing the cost of the best extension of
$(r_{n-3}=a, r_{n-2}=b)$ to the removed variables $r_n$ and
$r_{n-1}$. This process can be iterated (lines 4-8) to eliminate
variables up to $r_3$. Optimal values for variables $r_1$ and $r_2$
can be calculated using an exhaustive search (line 9). At this time,
the optimal cost can be calculated and the optimal values for
remaining variables can be set in increasing order using their
bucket and variables assigned beforehand (lines 11-14).

Note that the space complexity of the algorithm, when used as an
exact method, is $\Theta(n \times 2^{2n})$, due to the memory
required to store extensionally $n$ cost functions $g_i$, having
each $2^n \times 2^n$ entries. The time complexity is
$\Theta(n^2\times2^{3n})$ due to lines 4-8, as finding the minimum
of $2^n$ alternatives, being the computation of each one
$\Theta(n)$, has to be repeated $\Theta(n \times 2^{2n})$ times. On
the other hand, a basic search-based solution to the problem could
be implemented with worst case time complexity $\Theta(2^{(n^2)})$
and polynomial space. Observe that the time complexity of \BE\ is
therefore an exponential improvement over basic search algorithms,
although its high space complexity makes the approach impractical
for large instances.

\begin{figure}[h!]
\begin{algorithm}{Bucket Elimination for the \MDSLP}{}{}
     \LnNoNumber  {\bf function} BE($n,{\cal D}$)
     \LnNumber     {\bf for} $a, b \in {\cal D}$ {\bf do}
     \LnNumber     \> $g_n(a,b) := min_{c \in {\cal D}}\{f_{n-1}(a,b,c)+f_n(b,c)\}$
     \LnNumber     {\bf end for}
     \LnNumber     {\bf for} $i := n-1$ {\bf downto} $3$ {\bf do}
     \LnNumber     \> {\bf for} $a, b \in {\cal D}$ {\bf do}
     \LnNumber     \> \> $g_i(a,b) := min_{c \in {\cal D}}\{f_{i-1}(a,b,c)+g_{i+1}(b,c)\}$
     \LnNumber     \> {\bf end for}
     \LnNumber     {\bf end for}
     \LnNumber     $(r_1,r_2) := \mathrm{argmin}_{a,b \in {\cal D}}\{g_3(a,b)+f_1(a,b)\}$
     \LnNumber     $opt := g_3(r_1,r_2)+f_1(r_1,r_2)$
     \LnNumber     {\bf for} $i := 3$ {\bf to} $n-1$ {\bf do}
     \LnNumber     \> $r_i := \mathrm{argmin}_{c \in {\cal D}}\{f_{i-1}(r_{i-2},r_{i-1},c)+g_{i+1}(r_{i-1},c)\}$
     \LnNumber     {\bf end for}
     \LnNumber     $r_n := \mathrm{argmin}_{c \in {\cal D}}\{f_{n-1}(r_{n-2},r_{n-1},c)+f_{n}(r_{n-1},c)\}$
     \LnNumber    {\bf return} $(opt,(r_1,r_2,\dots,r_n))$
     \LnNoNumber  {\bf end function}
\end{algorithm}
\caption{Bucket elimination for the \MDSLP.}
\label{stilllife:fig:be:algorithm:mdslp}
\end{figure}

One interesting optimization, presented by
\citeA{DBLP:conf/cp/LarrosaM03}, allows reducing the complexity of
the algorithm. In the following, we assume that $n$ is even,
although a similar reasoning can be used if the size of the board is
odd. The optimization avoids the computation needed to eliminate
variables $r_1, r_2, \dots, r_{\frac{n}{2}-1}$, as a result of the
symmetry of the problem. In this way, the algorithm starts by
eliminating variables $r_n, r_{n-1}, \dots r_{\frac{n}{2}+2}$.
Observe that, at this point, cost functions $g_n, g_{n-1}, \dots,
g_{\frac{n}{2}+2}$ have been computed. At this point, the order to
eliminate remaining variables can be changed to $r_1, r_2, \dots,
r_{\frac{n}{2}-1}$. The elimination of $r_1$ would produce $g_1$
with scope $\{r_1,r_2\}$, but this computation can be avoided, as it
is the same to eliminate $r_1$ or to rotate the board by 180 degrees
and eliminate variable $r_n$, i.e.:
\begin{equation}
g_1(a,b) = g_n(\overline{b},\overline{a}),
\end{equation}
where $\overline{r}$ denotes the reflection value of the binary
string $r$. Moreover, if the board is vertically reflected, an
equivalent problem is obtained, so it follows that
\begin{equation}
g_n(\overline{b},\overline{a}) = g_n(b,a),
\end{equation}
and hence
\begin{equation}
g_1(a,b) = g_n(b,a).
\end{equation}
The optimized algorithm is obtained by applying the same reasoning
to the rest of the variables. \citeA{DBLP:conf/cp/LarrosaM03} and
\citeA{Larrosa05} have used this method\footnote{Actually, an
instance of the algorithm in which each variable is allowed to take
values in the whole computation domain. This is not our case as it
will be shown in next sections.} to solve the \MDSLP\ up to size 14.
The fourth column in Table~\ref{stilllife:table:Related:Work}
reproduces their results, obtained with a 2GHz Pentium IV machine
with 2Gb of memory. Notice the limitations of the approach: the
$n=15$ instance could not be solved due to space restraints. In
Section \ref{stilllife:sect:MA}, we will show how BE can be embedded
in a MA with reduced complexity in order to implement a smart
recombination operator.

\section[A Multi-Level Memetic/Exact Hybrid for the  MDSLP]{A Multi-Level Memetic/Exact Hybrid Algorithm for the  MDSLP}
\label{sect:multilevel} WCSPs are very amenable for being undertaken
with evolutionary metaheuristics. Obviously, the quality of the
results will greatly depend on how well knowledge of the problem is
incorporated into the search mechanism. Our final goal is to present
an algorithmic model based on the hybridization of MAs with exact
techniques at two levels: within the MA (as an embedded operator),
and outside it (in a cooperative model). Firstly, we will  focus in
the next subsection on the first level of hybridization, that
incorporates an exact technique (namely BE) within the MA as an
embedded recombination operator. Subsequently, we will proceed to a
second level of hybridization, in which the MA cooperates with a
branch-and-bound based beam search algorithm.

\subsection{A Memetic Algorithm with BE for the MDSLP}
\label{stilllife:sect:MA} In this subsection we describe a MA for
the \MDSLP\ that uses tabu search (TS) as a local search operator
and BE as an optimal recombination operator. Before detailing these
two components, let us describe the basic underlying evolutionary
algorithm (\EA).

\subsubsection{Representation and Fitness Calculation}
\label{stilllife:sect:fitness} The natural representation of
\MDSLP~solutions is their binary encoding. Accordingly, a
configuration for an $n \times n$ board will be represented as a
binary $n \times n$ matrix $r$. Clearly, infeasible solutions can be
represented, since not all such binary matrices will correspond to
stable patterns. One way to deal with this scenario is using
penalty-based fitness functions. To be precise, the fitness (to be
minimized) of a configuration $r$ is defined as:
\begin{eqnarray}
\label{stilllife:eq:fitness} f(r) = \sum_{i=1}^n\sum_{j=1}^n
(1-r_{ij})
      \mbox{}+  K\sum_{i=1}^{n+2}\sum_{j=1}^{n+2}
     \left[\embed{r}_{ij}\phi_1(\numNeighbors{\embed{r}}{i}{j}) +
     (1-\embed{r}_{ij})\phi_0(\numNeighbors{\embed{r}}{i}{j})
     \right].
\end{eqnarray}
Recall that stability is not only required within the $n \times n$
board, but also in its immediate neighborhood, and this is taken
into account by working with $\embed{r}$, the $(n+2)\times(n+2)$
binary matrix obtained by embedding $r$ in a frame of dead cells, as
defined in (\ref{stilllife:eq:embed}). $K$ is a constant,
$\numNeighbors{r}{i}{j}$ is the number of live neighbors of cell
$r_{ij}$, and $\phi_0$, $\phi_1:\mathbb{N}\longrightarrow\mathbb{N}$
are two functions (to be used with dead or alive cells
respectively), that take the number of alive neighbors of a cell,
and return a penalty depending on how many of them should be flipped
to have a stable configuration, defined as:
\begin{casos}
{\phi_0 (\eta)} \caso{0}{\eta \neq 3} \caso{K'+1}{\rm otherwise}
\otroCaso {\phi_1 (\eta)} \caso{0}{2 \leqslant \eta \leqslant 3}
\caso{K'+2-\eta}{\eta<2} \caso{K'+\eta-3}{\eta>3,}
\end{casos}
where $K'$ is another constant. The first double sum in
(\ref{stilllife:eq:fitness}) corresponds to the basic quality
measure for feasible solutions, i.e., its number of dead cells. With
respect to the last term, it represents the penalty for infeasible
solutions. The strength of penalization is controlled by constants
$K$ and $K'$. The values we have chosen for them ($K=n^2$ and
$K'=5n^2$) ensure that given any two solutions, the one that
violates less constraints is preferred; if two solutions violate the
same number of constraints, the one whose overall degree of
violation (i.e., distance to feasibility) is lower is preferred.
Finally, if the two solutions are feasible, the penalty term is null
and the solution with the higher number of live cells is better.

\subsubsection{A Local Improvement Strategy Based on Tabu Search}
\label{stilllife:sect:tabu:search}

The fitness function defined above provides a stratified notion of
gradient that can be exploited by a local search strategy. Moreover,
notice that the function is quite decomposable, since interactions
among variables are limited to adjacent cells in the board. Thus,
whenever a configuration is modified, the new fitness can be
computed only considering the cells located in adjacent positions to
changed cells. To be precise, assume that cell $(i,j)$ is modified
in solution $r$, resulting in solution $s$; the new fitness $f(s)$
can be computed as:
\begin{eqnarray}
    f(s)  =  f(r)
     + K\left[\Delta f_1(r_{ij},\numNeighbors{r}{i}{j}) +
     \sum_{c \in \neighborhood{r}{i}{j}}\Delta f_2(c,\funNumNeighbors(c),r_{ij})\right],
\end{eqnarray}
and functions $\Delta f_1$ and $\Delta f_2$ are defined as:
\begin{casos}
  {\Delta f_1(c,\eta)}
  \caso{0}{\eta = 2}
  \caso{(-1)^{(1-c)}\phi_0(\eta)}{\eta = 3}
  \caso{(-1)^c\phi_1(\eta)}{\rm otherwise}

  \otroCasoSinLeft{\Delta f_2(c',\eta,c)}{(1-c')\Delta f_{2,0}(\eta,c) + c'\Delta f_{2,1}(\eta,c)}

  \otroCaso{\Delta f_{2,0}(\eta,c)}
  \caso{K'+1}{\begin{array}[t]{l}
              (\eta = 2 \wedge c = 0)
               \vee(\eta = 4 \wedge c = 1)
              \end{array}}
  \caso{-(K'+1)}{\eta = 3}
  \caso{0}{\rm otherwise}

  \otroCaso{\Delta f_{2,1}(\eta,c)}
  \caso{K'+1}{\begin{array}[t]{l}
              (\eta = 2 \wedge c = 1)
              \vee(\eta = 3 \wedge c = 0)
              \end{array}}
  \caso{-(K'+1)}{\begin{array}[t]{l}
                 (\eta = 1 \wedge c = 0)
                  \vee(\eta = 4 \wedge c = 1)
                 \end{array}}
  \caso{1}{\begin{array}[t]{l}
           (\eta = 1 \wedge c = 1)
            \vee(\eta \geqslant 4 \wedge c = 0)
           \end{array}}
  \caso{-1}{\begin{array}[t]{l}
            (\eta = 0)
             \vee(\eta \geqslant 5 \wedge c = 1)
             \end{array}}
  \caso{0}{\rm otherwise.}
\end{casos}

Using this efficient fitness re-computation mechanism, our local
search strategy explores the neighborhood $N(r) = \left\{s~|~{\rm
Hamming}(r,s)=1\right\}$, i.e., the set of solutions obtained by
flipping exactly one cell in the configuration. This neighborhood
comprises $n^2$ configurations, and it is fully explored in order to
select the best neighbor. In order to escape from local optima, a
tabu-search scheme is used: up-hill moves are allowed, and after
flipping a cell, it is put in the tabu list for a number of
iterations (randomly drawn from $[n/2,3n/2]$ to hinder cycling in
the search). Thus, it cannot be modified in the subsequent
iterations unless the aspiration criterion is fulfilled. In this
case, the aspiration criterion is improving the best solution found
in that run of the local search strategy. The whole process is
repeated until a maximum number of iterations is reached, and the
best solution found is returned.

\subsubsection{Optimal recombination with \BE}
\label{stilllife:sect:Optimal:recombination:with:BE} Recall that the
fitness function that we have defined is able to evaluate any
representable configuration (feasible or not), and hence, the binary
representation used turns out to be freely manipulable. With this
setting, any standard recombination operator for binary strings
could be used in principle. For example, the two-dimensional version
of single-point crossover (2D-SPX), depicted in
Fig.~\ref{stilllife:fig:blind:recomb}, could be employed. Although
such a blind operator is feasible from a computational point of
view, it would perform poorly, as it would behave like a
macromutation operation. In order to achieve a sensible
recombination of information, we can resort to \BE.

Even though the performance of \BE\ as an exact method for the
\MDSLP\ was better than basic search-based approaches,  it was shown
in Section \ref{stilllife:sect::SLP:WCSP} that the corresponding
time and space complexity were still very high, making it unsuitable
for large instances. In the following, we explain how \BE\ can be
used to implement an intelligent recombination operator for the
\MDSLP. Such operator will explore the dynastic potential
\cite{Radcliffe_algebra_94b} (possible children) of the solutions
being recombined, providing the best solution that can be
constructed without introducing implicit mutation, i.e., exogenous
information (cf. \citeR{CT03:BnB-EA:APIN}). Moreover, we will show
that this operator is tractable from a computational point of view.

For this purpose, let  $x=(x_1,x_2,\cdots,x_n)$ and
$y=(y_1,y_2,\cdots,y_n)$ be two board configurations for an $n
\times n$ instance of the \MDSLP. Our operator will calculate the
best configuration that can be obtained by combining rows in $x$ and
$y$ without introducing information not present in any of the
parents. This can be achieved by restricting the domain of variables
in \BE\ to take values corresponding to the rows of the
configurations being recombined. Using the optimized version of the
BE algorithm, the recombination operator becomes
$\textnormal{BE-Opt}(n,\{x_1,x_2,\cdots,x_n,y_1,y_2,\cdots,y_n\})$,
so that the result returned by this invocation to the algorithm is
the best possible recombination.

\begin{figure}[h!]
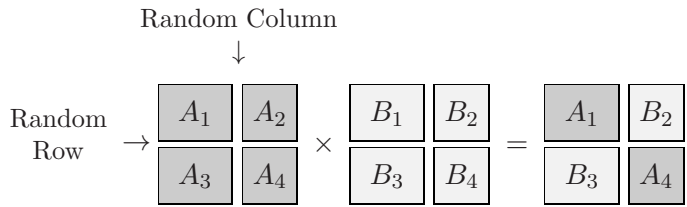

   \centering

   \definecolor{gray1}{gray}{0.8}
   \definecolor{gray2}{gray}{0.95}
   $$
   \begin{array}{l}
   \textnormal{\parbox{6.5cm}{\centering\small{Random Column\\$\downarrow$}}}\\
   \vspace*{-2mm}\\
   \parbox{1.7cm}{\vspace*{-3mm}\centering\small{ Random\ Row}}
   \parbox{0.5cm}{$ \rightarrow$}
   \hspace {-0.2cm}
   \begin{array}{c@{\ }c}
      \fcolorbox{black}{gray1}{\parbox{0.75cm}{\centering\vspace{1mm}$A_1$\vspace{1mm}}} & \fcolorbox{black}{gray1}{\parbox{0.5cm}{\centering\vspace{1mm}$A_2$\vspace{1mm}}} \\
   \   \vspace*{-0.4cm} \\
      \fcolorbox{black}{gray1}{\parbox{0.75cm}{\centering\vspace{1mm}$A_3$\vspace{1mm}}} & \fcolorbox{black}{gray1}{\parbox{0.5cm}{\centering\vspace{1mm}$A_4$\vspace{1mm}}} \\
   \end{array}
   \!\times\!
   \begin{array}{c@{\ }c}
      \fcolorbox{black}{gray2}{\parbox{0.75cm}{\centering\vspace{1mm}$B_1$\vspace{1mm}}} & \fcolorbox{black}{gray2}{\parbox{0.5cm}{\centering\vspace{1mm}$B_2$\vspace{1mm}}} \\
   \   \vspace*{-0.4cm} \\
      \fcolorbox{black}{gray2}{\parbox{0.75cm}{\centering\vspace{1mm}$B_3$\vspace{1mm}}} & \fcolorbox{black}{gray2}{\parbox{0.5cm}{\centering\vspace{1mm}$B_4$\vspace{1mm}}} \\
   \end{array}
   \!=\!
   \begin{array}{c@{\ }c}
      \fcolorbox{black}{gray1}{\parbox{0.75cm}{\centering\vspace{1mm}$A_1$\vspace{1mm}}} & \fcolorbox{black}{gray2}{\parbox{0.5cm}{\centering\vspace{1mm}$B_2$\vspace{1mm}}} \\
   \   \vspace*{-0.4cm} \\
      \fcolorbox{black}{gray2}{\parbox{0.75cm}{\centering\vspace{1mm}$B_3$\vspace{1mm}}} & \fcolorbox{black}{gray1}{\parbox{0.5cm}{\centering\vspace{1mm}$A_4$\vspace{1mm}}} \\
   \end{array}
   \end{array}
   $$

   \caption{Blind recombination operator for the \MDSLP.}
   \label{stilllife:fig:blind:recomb}
\end{figure}

In order to analyze the time complexity for this recombination
operator, the critical part of the algorithm is the execution of
lines 4-8 in Fig.~\ref{stilllife:fig:be:algorithm:mdslp}. In this
case, line 6 has complexity $O(n^2)$ (finding the minimum of at most
$2n$ alternatives, the computation of each being $\Theta(n)$). Line
6 has to be executed $n/2 \times 2n \times 2n$ times at most (recall
the optimization), making a global complexity of $O(n^5)$ =
$O(|x|^{2.5})$, where $|x|\in\Theta(n^2)$ is the size of solutions.
Notice also that the recombination procedure can be readily made to
further exploit the symmetry of the problem, extending variable
domains to column values in addition to row values. The complexity
bounds remain the same in this case.

One interesting property of the described operator is that it  can
be generalized to recombine any number of board configurations like
$\textnormal{BE-Opt}(n,\bigcup_{x \in S} \{ x_i\ |\ i \in \{1
\dotdot n\}\})$, where $S$ is a set comprising the solutions to be
recombined. In this situation, the time complexity is $O(k^3n^5)$
(line 6 is $O(kn^2)$, and it is executed $O(k^2n^3)$ times), where
$k=|S|$ is the number of configurations being recombined. This
multi-parental capability will be explored in the rest of the paper.

\subsubsection{Experimental Results}
\label{stilllife:sect:exp:res:MA}

In order to evaluate the usefulness of the described hybrid
recombination operator, a set of experiments for problem sizes from
$n=12$ up to $n=20$ has been realized (recall that optimal solutions
to the \MDSLP\ are known up to $n=20$). The experiments were
performed using a steady-state evolutionary algorithm
($popsize=100$, $p_m=1/n^2$, $p_X=0.9$, binary tournament
selection). With the aim of maintaining diversity, duplicated
individuals were not allowed in the population. Algorithms were run
until an optimal solution was found or a time limit was exceeded.
This time limit was set to 3 minutes for problem instances of size
12 and was gradually incremented by 60 seconds for each size
increment. For each algorithm and each instance size, 20 independent
executions were run. All the experiments in this paper have been
performed in a Pentium IV PC (2400MHz and 512MB of main memory)
under SuSE Linux.

The base algorithm used is a MA using 2D-SPX for recombination, and
endowed with tabu search for local improvement ($maxiter = n^2$).
This algorithm is termed \MATS, and has been shown to be capable of
finding feasible solutions systematically, solving to optimality
instances with $n<15$ (see \MATS~in
Fig.~\ref{stilllife:fig:PartialBE}). Although the performance of the
algorithm degrades for larger instances, it provides distributions
for the solutions whose average relative distance to the optimum is
less than 5.29\% in all cases. This contrasts with the case of plain
EAs, which are incapable of finding even a feasible solution in most
runs \cite{GallardoCottaFernandezEVOCOP06}.

\begin{figure}[h!]
   \centering \epsfig{file=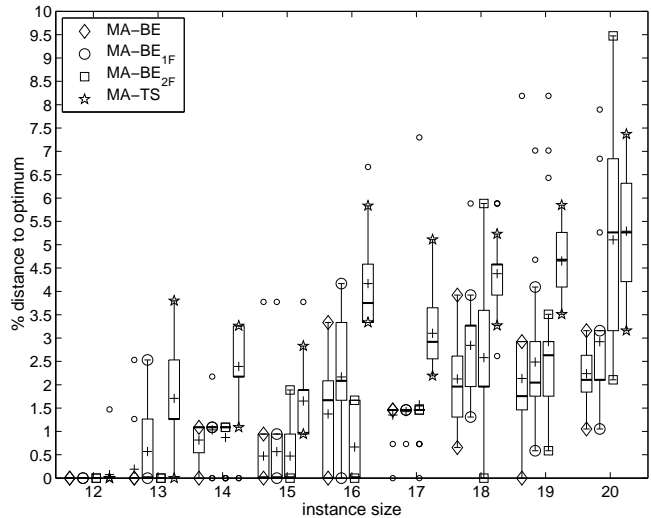,scale=.50} \caption[Relative
   distances to optimum for different algorithms for sizes ranging from
   12 up to 20.]{Relative
   distances to optimum for different algorithms for sizes ranging from
   12 up to 20. Each box summarizes 20 runs. In this and in all subsequent
   figures, boxes comprise the second and third quartiles of the
   distribution (i.e., the inner 50\%), an horizontal line marks the
   median, a plus sign indicates the mean, and circles indicate results
   further from the median than 1.5 times the interquartile-distance. } \label{stilllife:fig:PartialBE}
\end{figure}

\MATS~is firstly compared with MAs endowed with BE for performing
recombination. Since the use of BE for recombination has a higher
computational cost than a simple blind recombination, and there is
no guarantee that recombining two infeasible solutions will result
in a feasible solution, we have defined three variants of the MAs:
\begin{itemize}
\item In the first one,
denoted \MABE, BE is always used to perform recombination.
\item In the second one, termed \MABEONEF,
we require that at least one of the parents is feasible in order to
apply BE; otherwise blind recombination is used.
\item In the last one, identified as \MABEALLF,
we require the two parents to be feasible, thus being more
restrictive in the application of BE.
\end{itemize}
By evaluating these variants, we intend to explore the computational
tradeoffs involved in the application of BE as an embedded component
of the MA. For these algorithms, mutation was performed prior to
recombination in order to take advantage of good solutions provided
by BE. Fig.~\ref{stilllife:fig:PartialBE}  shows the empirical
performance of the different algorithms evaluated (as the relative
distance to the optimum). Results show that \MABE~improves
significantly over \MATS~and can find better solutions.
\MABEALLF~can find slightly better solutions than \MABE~on smaller
instances ($n \in \{13,15,16\}$), but on larger instances the winner
is \MABE. It seems that the effort saved not recombining unfeasible
solutions does not further improve the performance of the algorithm.
Note also that, for larger instances, \MABEONEF~is better than
\MABEALLF. This correlates well with the fact that BE is used more
frequently in the former than in the latter.

As mentioned in Section
\ref{stilllife:sect:Optimal:recombination:with:BE}, the optimal
recombination scheme we use can be readily extended to multi-parent
recombination \cite{Eiben_et_al_Multiparent_Recombination}: an
arbitrary number of solutions can contribute their constituent rows
for constructing a new solution. Additional experiments were done to
explore the effect of this capability of \MABE.
Fig.~\ref{stilllife:fig:MABE} shows the results obtained by
\MABE~for a different number of parents being recombined (arities 2,
4, 8 and 16). For $arity=2$, the algorithm was able to find the
optimum solution for all instances except for $n=18$ and $n=20$ (the
relative distance to the optimum for the best solution found is less
than 1.04\% in these cases). Executions with $arity=4$ cannot find
optimum solutions for the remaining instances, but note that the
distribution improves in some cases. Clearly, the performance of the
algorithm deteriorates when combining more than 4 parents due to the
higher computational cost. Variable clustering could be used to
alleviate this higher computational cost, but this results in
performance degradation since the more coarse granularity of the
information pieces hinders information mixing
\cite{CT00a,DBLP:conf/ppsn/CottaDFH06}.

\begin{figure}[h!]
   \centering \epsfig{file=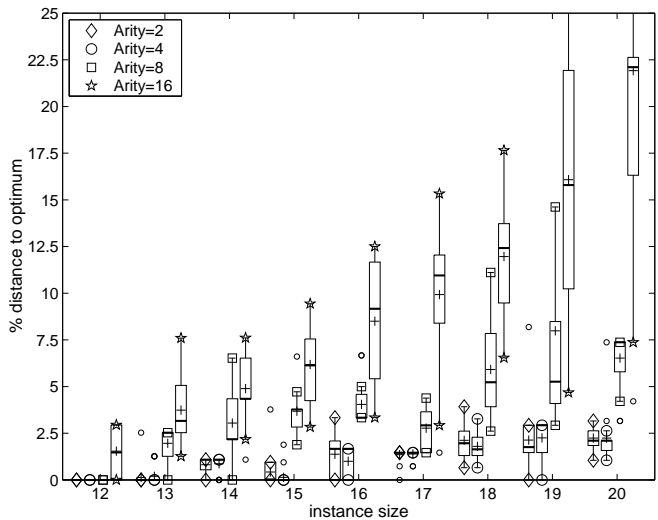,scale=.50}
   \caption[Relative distances to optimum for different
   arities for \MABE~for sizes ranging from 12 up to 20.]{Relative distances to optimum for different
   arities for \MABE~for sizes ranging from 12 up to 20.
   Each box summarizes 20 runs.}
\label{stilllife:fig:MABE}
\end{figure}

\subsection{A Beam Search/MA Hybrid Algorithm}
\label{stilllife:sect:Hybrid}

\def\xx{\hspace{0.1ex}}
\def\mm{{\xx+\kern-1.0ex+\xx}}

In this subsection, we describe a hybrid tree search/memetic
algorithm for the \MDSLP. This algorithm combines, in a
collaborative way, a BS algorithm and a MA. As noted before, BS
works by extending in parallel a set of different partial solutions
in several possible ways, and thus can be used to provide promising
partial solutions to a population based search method such as a MA.
The goal is to exploit the capability of BS for identifying probably
good regions of the search space, and the strength of the MA for
exploring these, synergistically combining these two different
approaches.

The proposed hybrid algorithm, that  executes BS and the MA in an
interleaved way, is depicted in
Fig.~\ref{stilllife:fig:hybrid:algorithm}. In the pseudo-code, a
(possible partial) solution for an $n \times n$ instance is
represented by a vector of rows $s=(r_1, r_2, \dots, r_i),\ i
\leqslant n$, where rows are encoded as binary strings, $s \cdot
(r_i = v)$ stands for the extension of partial solution $s$ by
assigning value $v$ to its $i$-th row, and $\overline{v}$ denotes
the reflection value of the binary string $v$. The hybrid algorithm,
Hybrid($n,k_{bw},k_{MA}$), constructs a search tree, such that its
leaves consist of all possible board configurations of size $n
\times n$ that can be generated using solely symmetric rows (this
symmetry constraint was imposed to keep the branching factor,
$k_{ext}$, of the BS at a manageable level for the range of instance
sizes considered), and internal nodes at level $i$ represent
partially specified (up to the $i$-th row) board configurations.
This tree is incompletely traversed in a breadth first way using a
BS algorithm with beam width $k_{bw}$ (i.e., maintaining only the
best $k_{bw}$ nodes at each level of the tree). For the beam
selection (line 10), a simple quality measure is defined for partial
solutions, whose value is either $\infty$ if the partial
configuration is unstable, or its number of dead cells otherwise.
The algorithm starts (line 2) with a totally unspecified solution
(i.e., a solution with 0 rows). Initially, only the BS part of the
algorithm is executed. During each iteration of the BS (lines 3-17),
a new row is added to every solution in the beam (line 7). The
interleaved execution of the MA starts only when  partial solutions
in the beam have at least $k_{MA}$ rows (line 11). For each
iteration of the BS, the best $popsize$ solutions in the beam are
selected (using the quality measure described above) to initialize
the population of the MA (line 12). Since these are partial
solutions, they must be first converted into full solutions, e.g.,
by completing remaining rows randomly. After running the MA, its
solution is used to update the incumbent solution ($sol$), and this
process is repeated until the search tree is exhausted.

\begin{figure}[h!]
     \begin{algorithm}{Hybrid algorithm for the \MDSLP}{}{}
     \LnNoNumber {\bf function} Hybrid ($n,k_{bw},k_{MA}$)
     \LnNumber  $sol := \infty$
     \LnNumber  $\beam := \{\ ()\ \}$
     \LnNumber {\bf for} $i := 1$ {\bf to} $n$ {\bf do}
     \LnNumber \> $\beam' := \{\}$
     \LnNumber \> {\bf for} $s \in \beam$ {\bf do}
     \LnNumber \> \> {\bf for} $r := 0$ {\bf to} $2^{\lceil n/2 \rceil}-1$ {\bf do}
     \LnNumber \> \> \>  $\beam' := \beam'\ \cup$ $\{ s \cdot (r_i = r$ {\bf or} $\overline{r}) \}$
     \LnNumber \> \> {\bf end for}
     \LnNumber \> {\bf end for}
     \LnNumber \> $\beam$\ := {\bf select} best $k_{bw}$  nodes from $\beam'$
     \LnNumber \> {\bf if} ($i \geq k_{MA}$) {\bf then}
     \LnNumber \> \> {\bf initialize} MA population  with best $popsize$ nodes from  $\beam'$
     \LnNumber \> \> {\bf run} MA
     \LnNumber \> \> $sol :=$ {\bf min} $(sol, \mathrm{MA\ solution})$
     \LnNumber \> {\bf end if}
     \LnNumber {\bf end for}
     \LnNumber {\bf return} $sol$
     \LnNoNumber {\bf end function}
     \end{algorithm}
     \caption{Hybrid algorithm for the \MDSLP.}
     \label{stilllife:fig:hybrid:algorithm}
\end{figure}

\subsubsection{Experimental Results}

\begin{figure}[h!]
\centering $
\renewcommand{\arraystretch}{0}%
\begin{array}{@{}r@{}}
    \epsfig{file=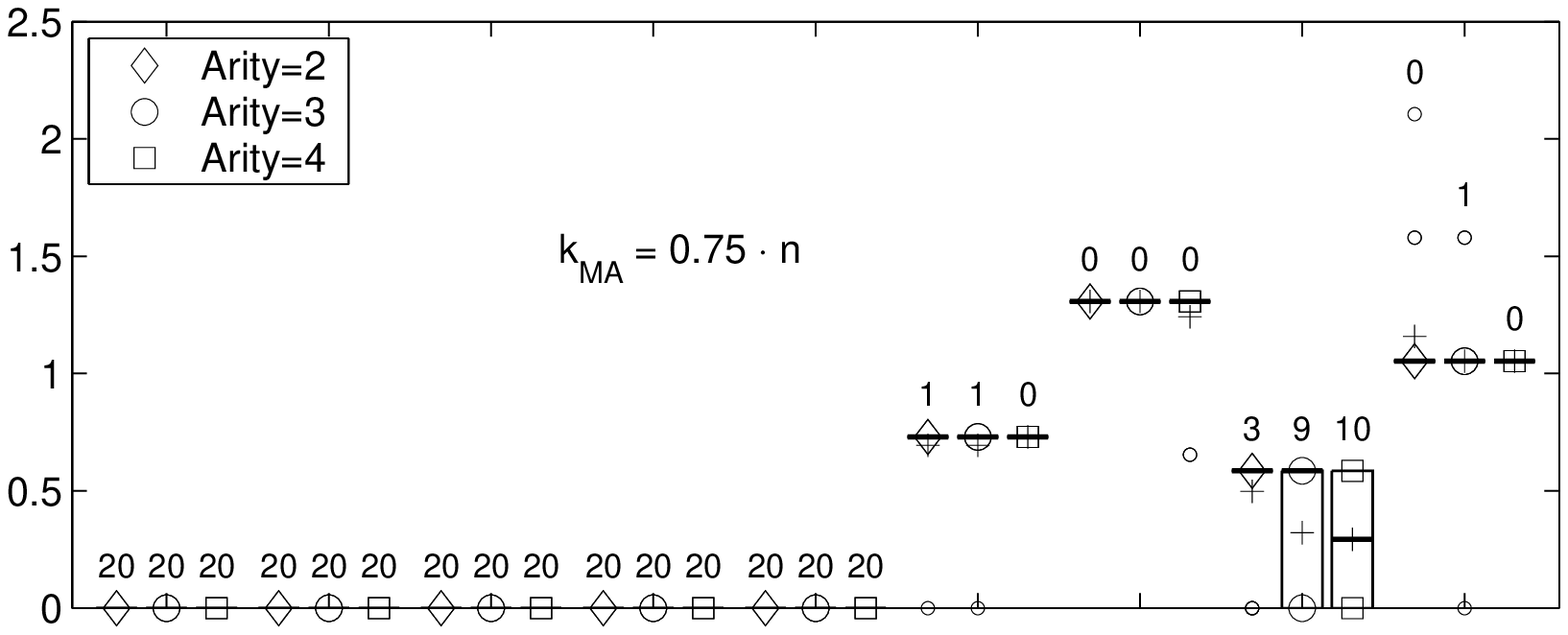,scale=.50}\\
    \epsfig{file=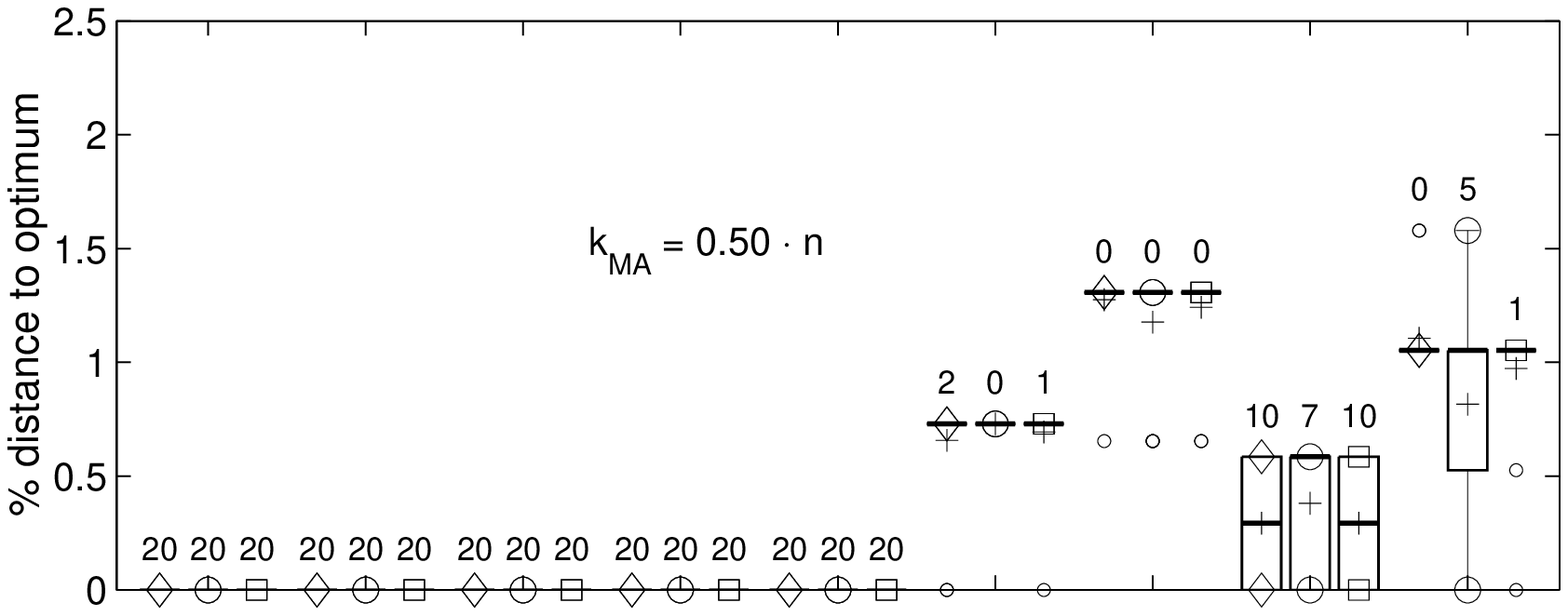,scale=.50}\\
    \epsfig{file=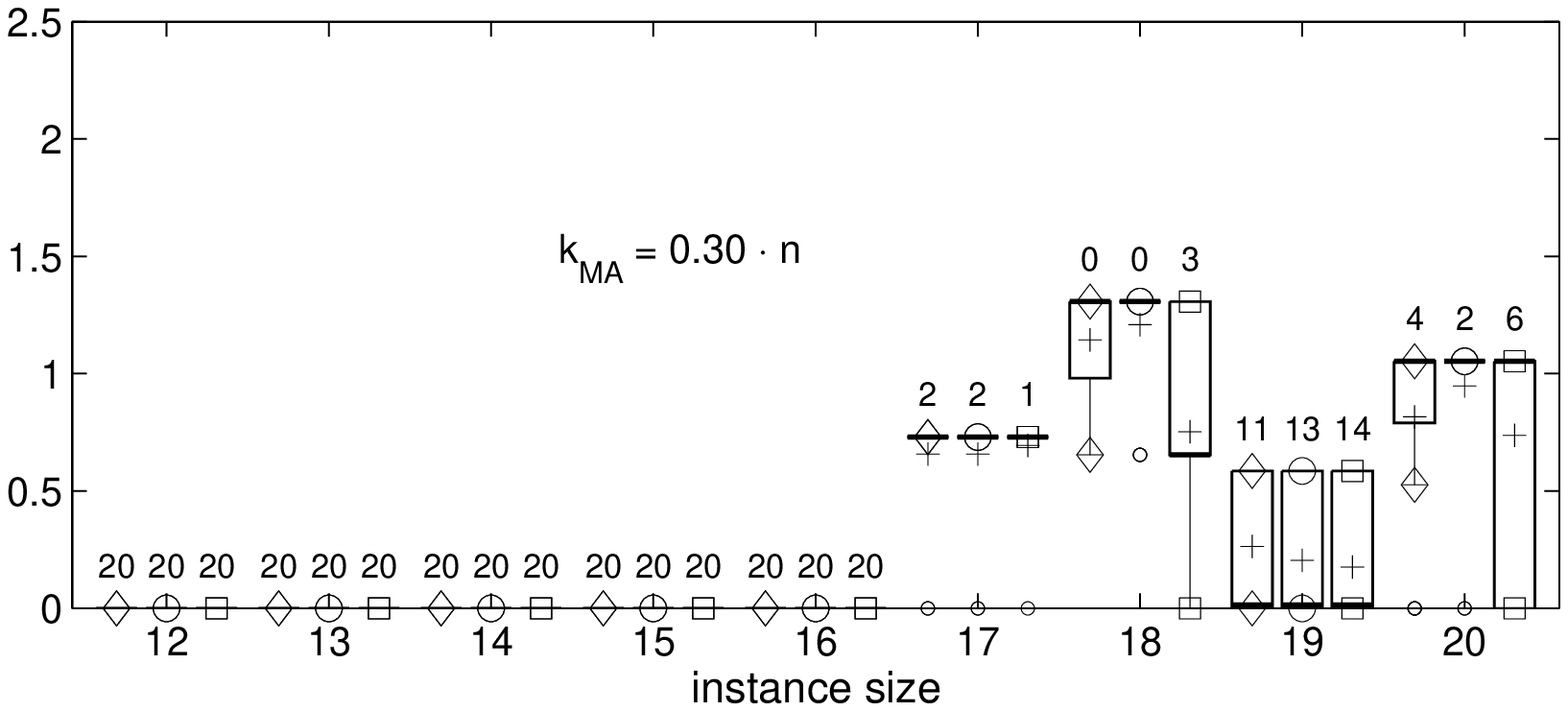,scale=.50}
\end{array}%
\renewcommand{\arraystretch}{0}%
$ \caption[Relative distances to optimum for different arities for
{\protect\linebreak[4]} \BSMABE~and $K_{MA} \in \{0.3 \cdot n, 0.5
\cdot n, 0.75 \cdot n\}$, for sizes ranging from 12 up to
20.]{Relative distances to optimum for different arities for
\BSMABE~and $K_{MA} \in \{0.3 \cdot n, 0.5 \cdot n, 0.75 \cdot n\}$,
for sizes ranging from 12 up to 20. Each box summarizes 20 runs. The
numbers above each box indicate how many times the optimal solution
was found.} \label{stilllife:fig:BS-MA-BE-kMA}
\end{figure}

Experiments were conducted to evaluate the hybrid algorithm
(\BSMABE). The methodology was the same as Section
\ref{stilllife:sect:exp:res:MA} (20 executions were performed for
each algorithm and instance size), but arities for the MA where in
$\{2,3,4\}$. The setting of parameters was $k_{bw}=2000$
(preliminary tests indicated that this value was reasonable), and
$k_{MA} \in \{0.3 \cdot n, 0.5 \cdot n, 0.75 \cdot n\}$, i.e., the
best 2000 nodes were kept on each level of the BS algorithm, and
30\%, 50\% or 75\% of the levels of the BS tree were initially
descended before starting to run the MA. With respect to termination
conditions, each execution of the MA within the hybrid algorithm
consisted of 1000 generations, and no time limits were imposed for
the hybrid algorithms that were run for $n$ iterations of the BS.

\begin{figure}[h!]
\centering $
\renewcommand{\arraystretch}{0}%
\begin{array}{@{}r@{}}
\epsfig{file=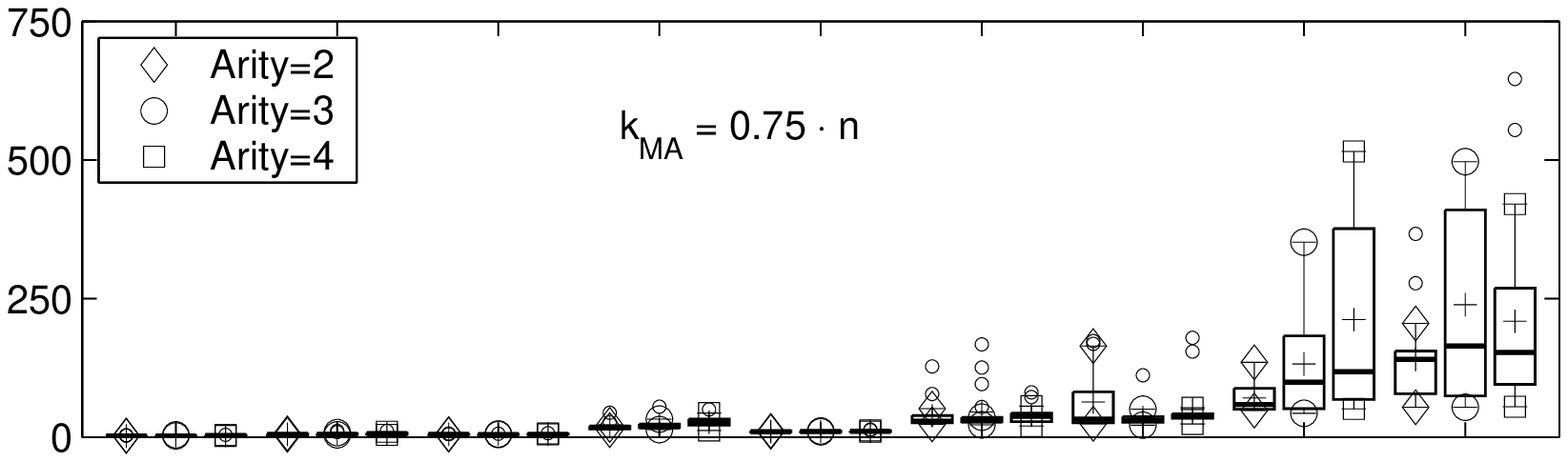,scale=.45}\\
\epsfig{file=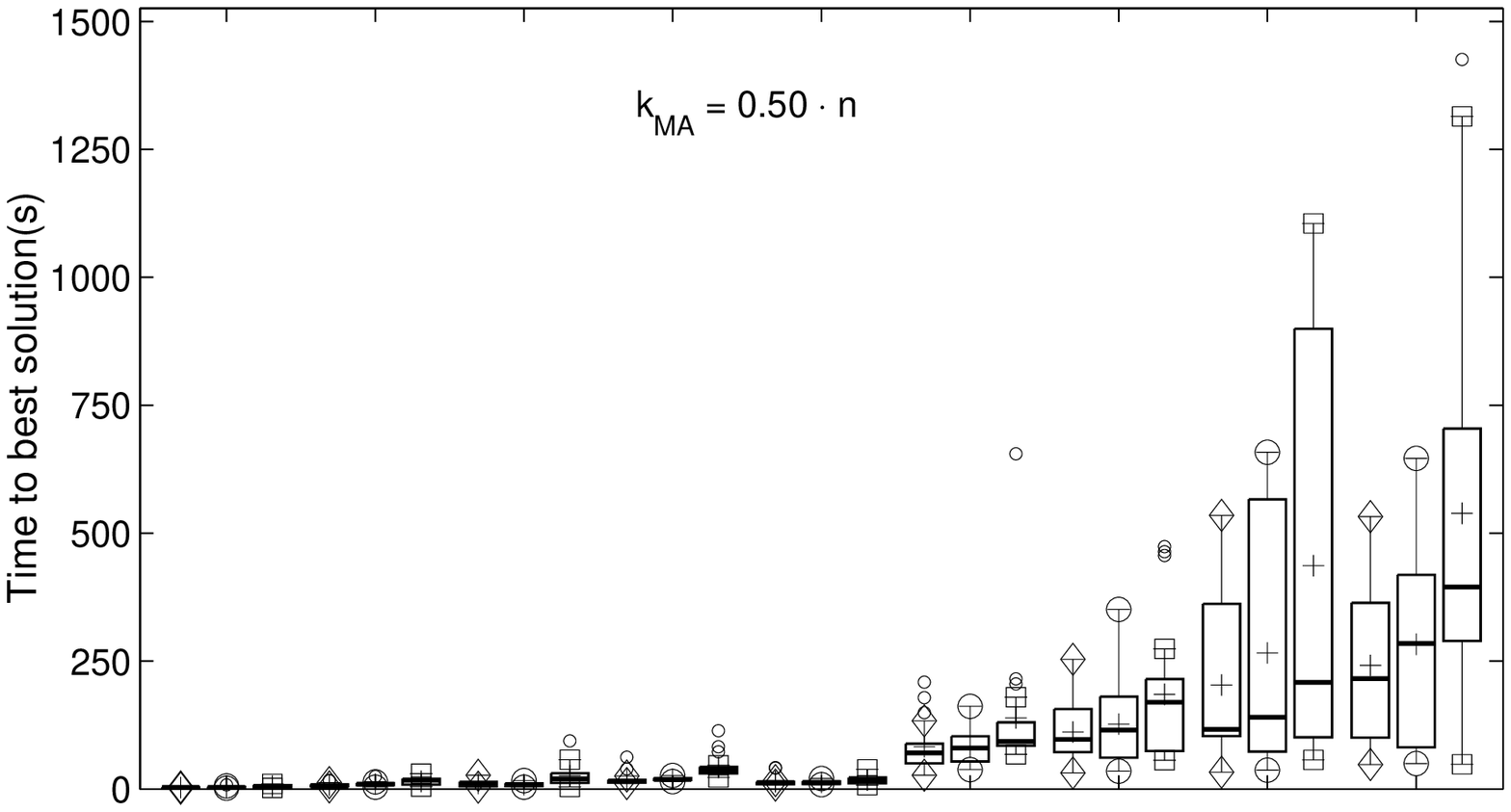,scale=.45}\\
\epsfig{file=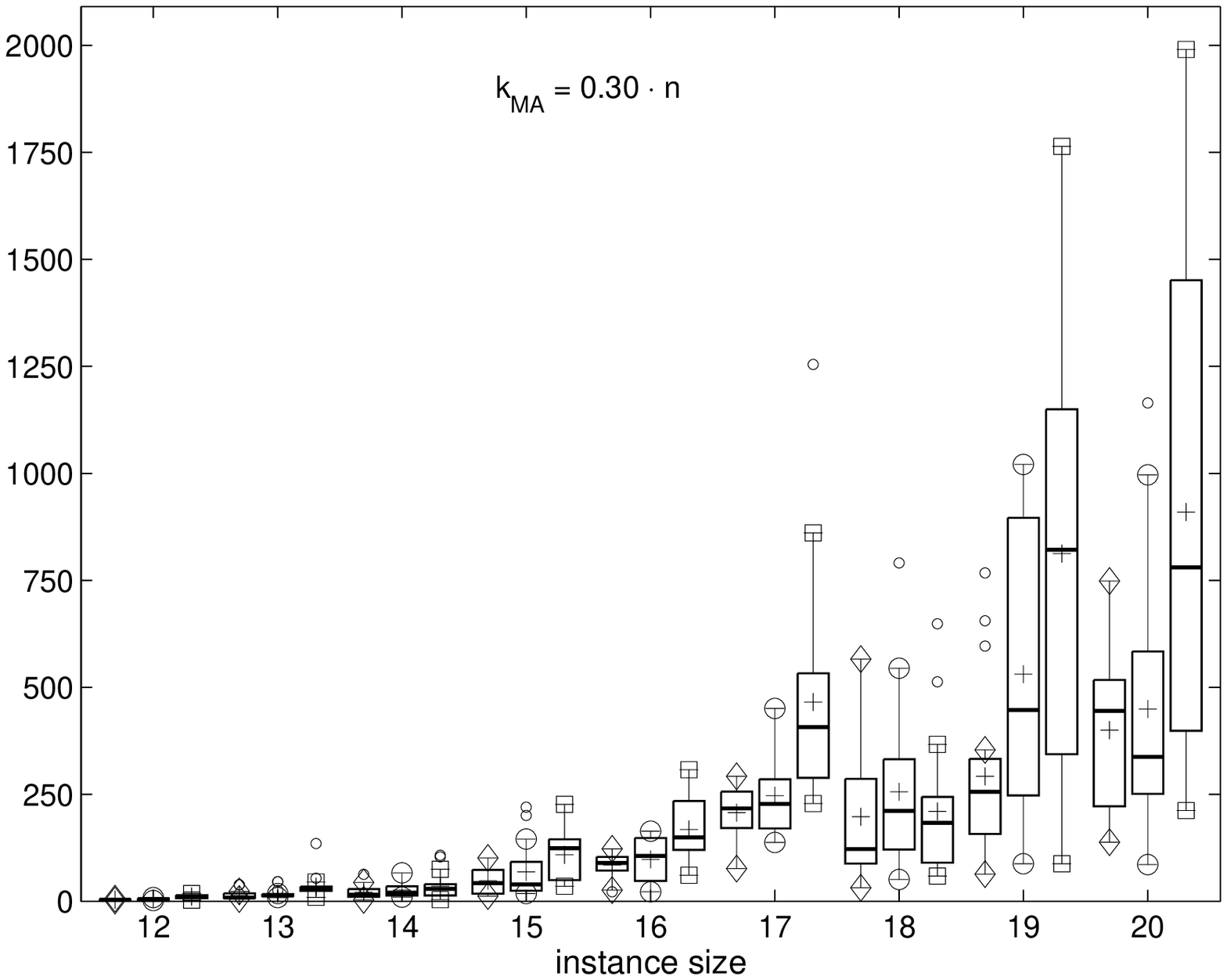,scale=.45}
\end{array}%
\renewcommand{\arraystretch}{0}%
$ \caption[Time  to best solution for different arities for
\BSMABE~and {\protect\linebreak[4]}$K_{MA} \in \{0.3 \cdot n, 0.5
\cdot n, 0.75 \cdot n\}$, for sizes ranging from 12 up to 20.]{Time
(in seconds) to best solution for different arities for \BSMABE~and
$K_{MA} \in \{0.3 \cdot n, 0.5 \cdot n, 0.75 \cdot n\}$, for sizes
ranging from 12 up to 20. Each box summarizes 20 runs.}
\label{stilllife:fig:BS-MA-BE-kMA-Time}
\end{figure}

Fig.~\ref{stilllife:fig:BS-MA-BE-kMA}  shows the results for
different values of parameter $k_{MA}$. In order to better compare
the distributions, the number of optimal solutions obtained by each
algorithm (out of 20 executions) is shown above each box plot. For
$k_{MA} = 0.3 \cdot n$, the performance of the resulting algorithm
improves significantly over the original MA. Note that \BSMABE,
using an arity of 2 parents, is able to find the optimum for all
cases except for $n=18$ (this instance is solved with $arity=4$).
All distributions for different instance sizes are significantly
improved. For $n<17$ and $arity \in \{2,3,4\}$, the algorithm
consistently finds  the optimum in all runs. For other instances,
the solution provided by the algorithm is always within a 1.05\% of
the optimum, except for $n=18$, for which the relative distance to
the optimum for the worst solution is 1.3\%. The other two charts
show that, in general, the performance of the algorithm deteriorates
with increasing values of the $k_{MA}$ parameter. This may be due to
the low quality of the bounds used in the BS part.

Regarding execution times,
Fig.~\ref{stilllife:fig:BS-MA-BE-kMA-Time} shows the distributions
for the time (in seconds) to reach the best solution needed by the
algorithms. Although \BSMABE~requires more time than \MABE, the
time needed remains reasonable for these instances, and is always
less than 2000 seconds. Note also how the execution time increases
with the arity, as more time is needed by the MA to perform BE in
the crossover operator. On the other hand, execution time
decreases for larger values of $k_{MA}$ as the number of
executions of the MA decreases, although, as we have already
remarked, the quality of the solutions worsens.

\begin{figure}[h!]
\centering \epsfig{file=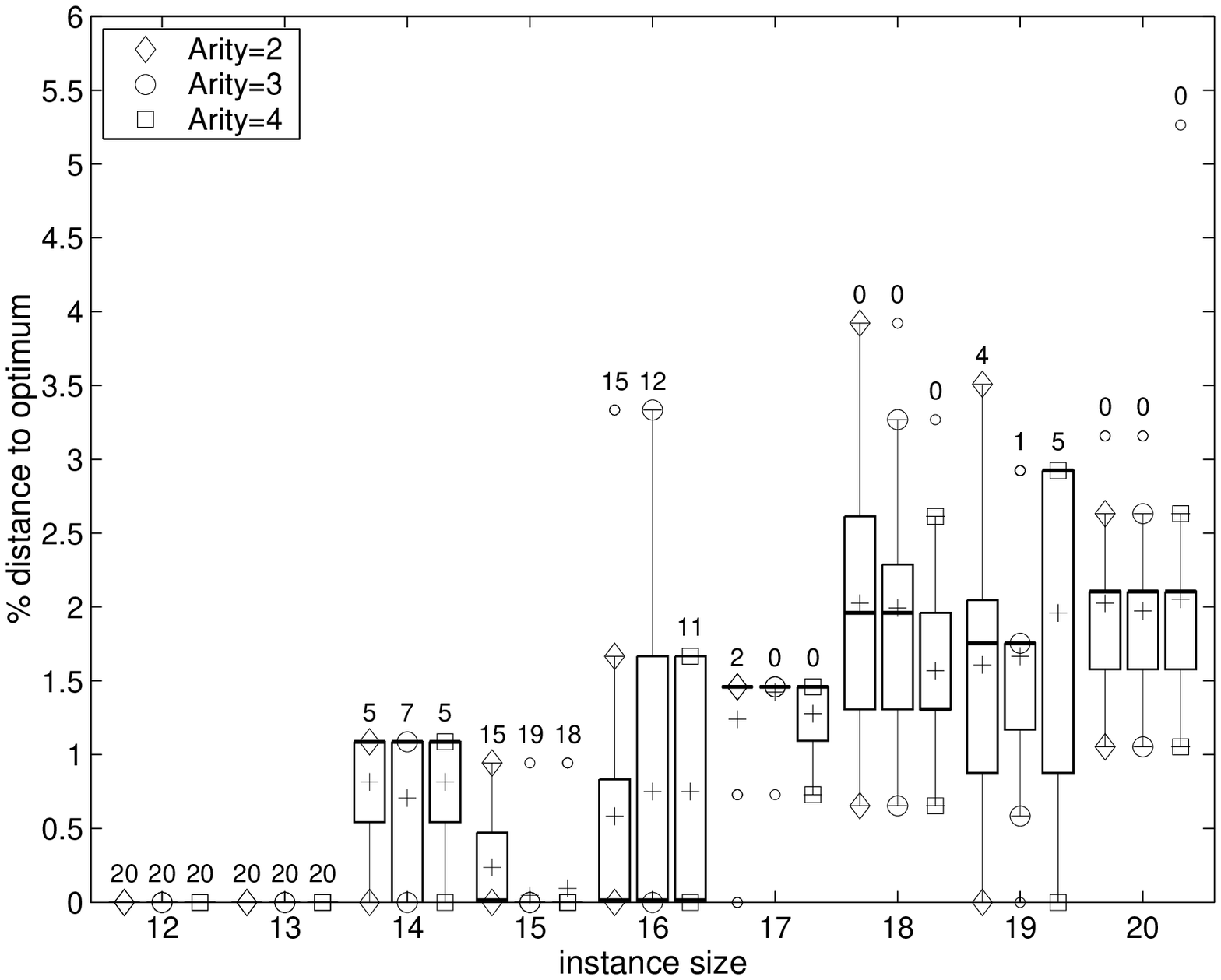,scale=.50}
\caption[Relative distances to optimum for different arities for
\MABE~executed for 2800 seconds, for sizes ranging from 12 up to
20.]{Relative distances to optimum for different arities for
\MABE~executed for 2800 seconds, for sizes ranging from 12 up to 20.
Each box summarizes 20 runs.} \label{stilllife:fig:MA-BE-TIME2800}
\end{figure}

To verify that the improved results of the hybrid algorithm were not
only a consequence of the extended execution times, experiments for
\MABE~were repeated with an increased time limit of 2800 seconds for
each execution independently of the instance size. The results of
these experiments are shown in
Fig.~\ref{stilllife:fig:MA-BE-TIME2800}. Clearly, the performance of
\MABE~does not improve dramatically, and this provides evidence on
the synergetic cooperation of BS and MA achieved by the hybrid
algorithm.

\section{A New Hybrid Algorithm Based on Mini-Buckets}
\label{sect:new:hybrid} In this section we present a novel hybrid
algorithm based on the algorithm described in
Section~\ref{stilllife:sect:Hybrid}. This algorithm exploits the
technique of Mini-buckets that is explained in the following.

\subsection{Mini-Buckets}
The main drawback of BE is that it requires exponential space to
store functions extensionally. When this complexity is too high, the
solution can be approximated using the technique of mini-buckets
(MB) presented by \citeA{dechter97minibuckets} (see also
\citeR{dechter03minibuckets}). Recall that, in order to eliminate
variable $x_i$, with its corresponding bucket $B_i=\{f_{i_1}, \dots
f_{i_m}\}$, BE
 calculates a new cost function
\begin{equation}
g_i = (\sum_{f \in B_i} f) \Downarrow x_i
\end{equation}
whose time and space complexity increases with the arity of $g_i$,
i.e., with the arity of the set $\bigcup_{f\in B_i} var(f) -
\{x_i\}$. This complexity can be decreased by approximating the
function $g_i$ with a set of smaller-arity functions. The basic
idea is to partition bucket $B_i$ into $k$ so called mini-buckets
$B_{i_1}, \dots, B_{i_k}$, such that the number of variables in
the scope of each $B_{i_j}$ is bounded by a parameter. Afterwards,
a set of $k$ cost functions with the reduced arity sought can be
defined as
\begin{equation}
g_{i_j} = (\sum_{f \in B_{i_j}} f) \Downarrow x_i, j=1\dots k,
\end{equation}
and the required approximation to $g_i$ can be computed as their
sum:
\begin{equation}
g'_i = \sum_{j=1}^{k} g_{i_j} = \sum_{j=1}^{k}\ \big( (\sum_{f \in
B_{i_j}} f) \Downarrow x_i \big)
\end{equation}
Note that the minimization computed in $g_i$ by the $\Downarrow$
operator has been migrated inside the sum. Since, in general, for
any two non-negative functions $f_1(x)$ and $f_2(x)$, $min_x
(f_1(x) + f_2(x)) \geq min_x f_1(x) + min_x f_2(x)$, the following
inequality holds
\begin{equation}
\overbrace{(\sum_{f \in B_i} f) \Downarrow x_i}^{g_i} \geq
\overbrace{\sum_{j=1}^{k} \big( (\sum_{f \in B_{i_j}} f)
\Downarrow x_i \big)}^{g'_{i}}
\end{equation}
and, thus  $g'_i$ is a lower bound on $g_i$. Therefore, if variable
elimination is performed using approximated cost functions, it
provides a lower bound for the optimal cost requiring less
computation than BE. Notice that the described approach provides a
family of under-estimating heuristic functions whose complexity and
accuracy is parameterized by the maximum number of variables allowed
in each mini-bucket.

\subsection{Improving the Lower Bound Using Mini-Buckets}
 The simple quality measure for beam selection used in
the algorithm in section~\ref{stilllife:sect:Hybrid} depends solely
on the part of the solution that is already constructed. In this
section, we will use the MB technique to compute a tight, yet
computationally inexpensive, lower bound for the remanning part of
the configuration with the aim of improving the performance of the
BS part of the hybrid algorithm.

For this purpose, let us note that the \MDSLP\ for an $n \times n$
board can be formulated as an alternative WCSP, if we associate a
different variable $x_{ij}$ for each cell $(i,j)$ on the board. With
this formulation, there are $n^2$ cost functions $f_{ij}, 1
\leqslant i,j \leqslant n$. The scope of function $f_{ij}$ is
$x_{ij}$ and all its neighborhood, and it returns $\infty$ if the
cell $(i,j$) is unstable, 1 if cell $(i,j)$ is dead, and 0
otherwise. The following objective function
\begin{equation}
F=\sum_{i=1}^{n}\sum_{j=1}^{n} f_{ij}
\end{equation}
has to be minimized.

Note that the original formulation, introduced in
Section~\ref{stilllife:sect::SLP:WCSP:modeling}, can be obtained
from the present one by clustering all cost functions corresponding
to row $r_i$ into a single cost function $f_i$. In the same way, let
us cluster cost functions for the $i$-th row into $M$  cost
functions, $f_i^1, f_i^2, \dots, f_i^M$ of roughly the same arity
($\approx n/M$), each one evaluating respectively one of the $M$
segments of the row. To be precise,
\begin{equation}
f_i^m=\sum_{j=1+\sum_{k=1}^{m-1}w_k}^{\sum_{k=1}^{m}w_k} f_{ij}, \ \
\ \ 1 \leqslant m \leqslant M
\end{equation}
where $w_n, 1 \leqslant n \leqslant M$, stands for the number of
variables of each segment.

Using this formulation, BE would perform the elimination of all
variables corresponding to the last row by computing a new $g_n$
cost function as
\begin{eqnarray}
g_n &=& (\sum_{m=1}^{M}f_{n-1}^m + \sum_{m=1}^{M}f_{n}^m) \Downarrow
\{x_{n1}, x_{n2}, \dots x_{nn}\},
\end{eqnarray}
whose bucket is $B_n =
\{f_{n-1}^1,f_{n-1}^2,\dots,f_{n-1}^M,f_n^1,f_n^2,\dots,f_n^M\}$.
Applying
 mini-buckets, $B_n$ can be partitioned into $M$ buckets: $B_n^m=\{f_{n-1}^m,f_{n}^m\}, 1\leq m \leq M$,
 and
a set of $M$ cost functions to approximate $g_n$ with reduced arity
can be calculated as:
\begin{eqnarray}
g_n^m &=& (f_{n-1}^m + f_{n}^m ) \Downarrow x_i^m,  \ \ \ \ 1 \leq m
\leq M
\end{eqnarray}
where
\begin{eqnarray}
x_i^1 &=&\{x_{i1},x_{i2},\dots ,x_{i(1+w_1)}\}\\
x_i^m &=&\{x_{i(\sum_{j=1}^{m-1}w_j)}, x_{i(1+\sum_{j=1}^{m-1}w_j)},
\dots, x_{i(1+\sum_{j=1}^{m}w_j)}\}, \ \ \ \ 1 < m
< M\\
x_i^M &=&\{x_{i(\sum_{j=1}^{M-1}w_j)},
x_{i(1+\sum_{j=1}^{M-1}w_j)},\dots,x_{in}\}.
\end{eqnarray}
In this way, the number of variables in each meta-variable $x_i^m,\
1 \leq m \leq M$, is $n/M$ approximately. Because the scopes of
$g_n^m,\ 1 \leq m \leq M$, are $\{x_{n-2}^m, x_{n-1}^m\}$, their
arities are approximately $1/M$  of the arity of $g_n$. The rest of
the rows of the board can be processed in a similar way.

Cost functions computed by the function MB can be used to estimate a
tight lower bound for a partial solution during the execution of the
hybrid algorithm as follows: let $s=(r_1,r_2,\dots,r_{k})$ be a
partial solution with $k$ rows for an $n \times n$ instance of the
\MDSLP.
 As defined, $g_{k+1}^1(r_{k-1}^1,r_{k}^1)$ returns the cost of the best extension to partial solution $s$
that can be attained in rows $k$ to $n$, considering only the first
column. In a similar manner, $g_{k+1}^m, 1 < m \leq n$  can be used
to estimate the best extension considering only columns 2 to $n$
respectively. Hence, a lower bound for a partial solution can be
computed as:
\begin{eqnarray}
lb(r_1,r_2,\dots,r_{k}) =  \sum_{i=1}^{k-1}\sum_{j=1}^{n} f_{ij} +
\sum_{m=1}^{n} g_{k+1}^m(r_{k-1}^m,r_{k}^m),
\end{eqnarray}
where the first sum corresponds to the part of the solution already
assigned. This bound can be used to rank nodes for beam selection
and the initialization of the MA population.

In the following subsection, we have experimented with setting
$M=3$, so that
\begin{casos}
{(w_1,w_2,w_3)} \caso{(n/3,n/3,n/3)}{n \bmod 3 =0} \caso{(\lfloor
n/3 \rfloor, \lceil n/3 \rceil,  \lfloor n/3 \rfloor)}{
    n \bmod 3 = 1} \caso{(\lceil n/3 \rceil,\lfloor n/3
\rfloor,\lceil n/3 \rceil)}{     n \bmod 3 = 2.}
\end{casos}
Observe that, for these settings, the space complexity of function
MB is $O(n \times 2^{2 (\lceil \frac{n}{3} \rceil+2)})$, whereas its
time complexity is $O(n^2 \times 2^{3 (\lceil \frac{n}{3}
\rceil+2)})$. When this complexity is still too high,
 the approach described in this subsection can be utilized
 to reduce it further,
considering more than three clustered cost functions for each row of
variables, although the resulting bounds would be less tight.

\subsubsection{Experimental Results}

\begin{figure}[h!]
\centering $
\renewcommand{\arraystretch}{0}%
\begin{array}{@{}r@{}}
    \epsfig{file=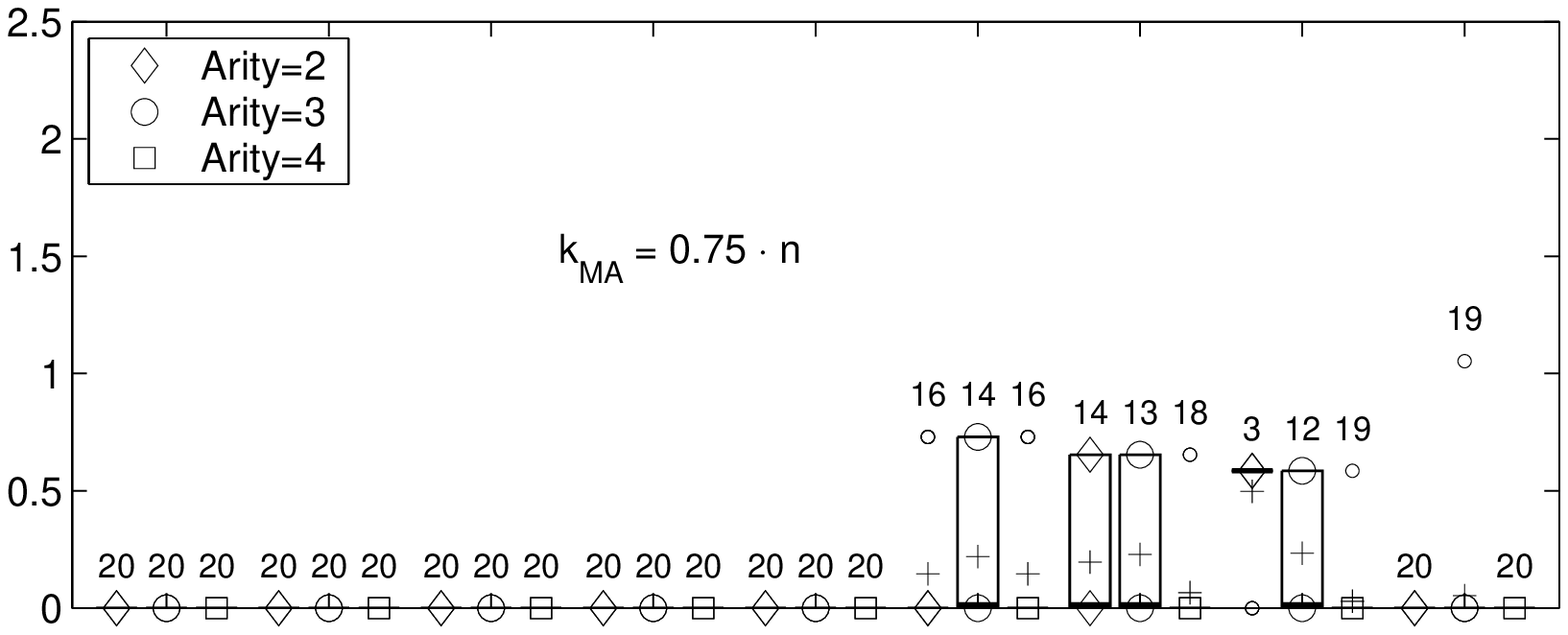,scale=.50}\\
    \epsfig{file=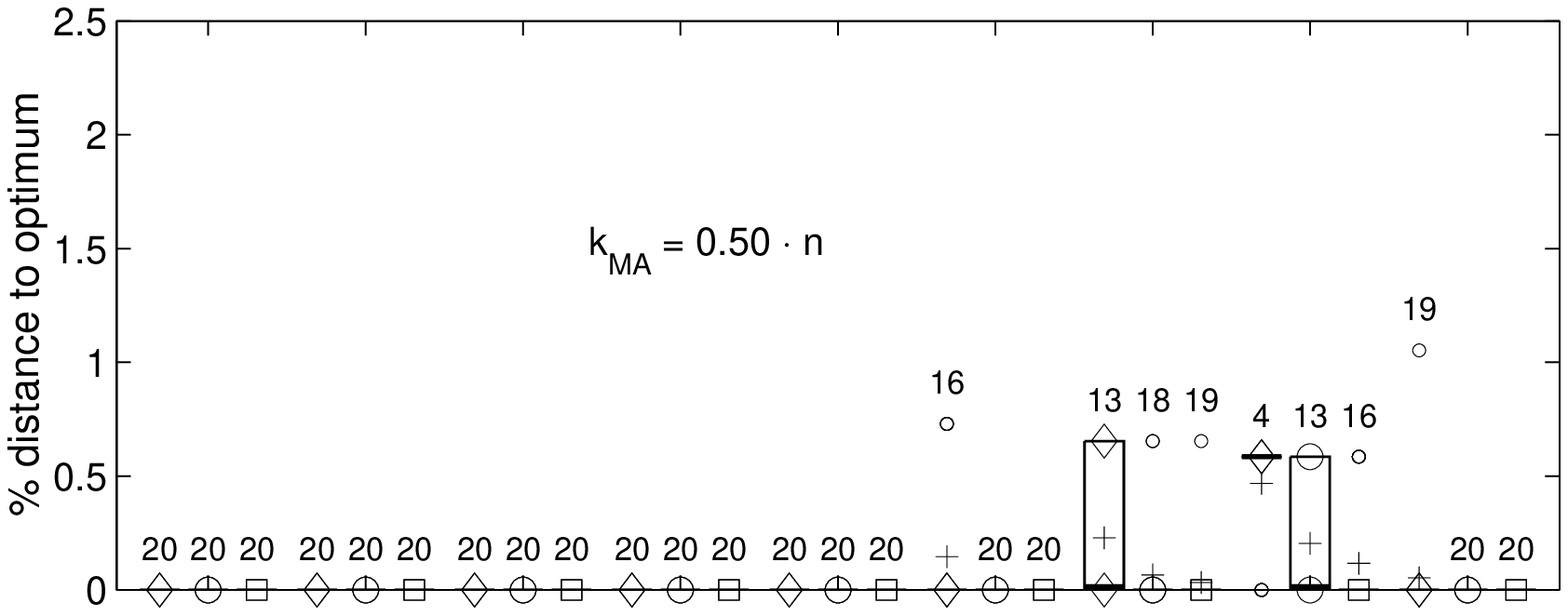,scale=.50}\\
    \epsfig{file=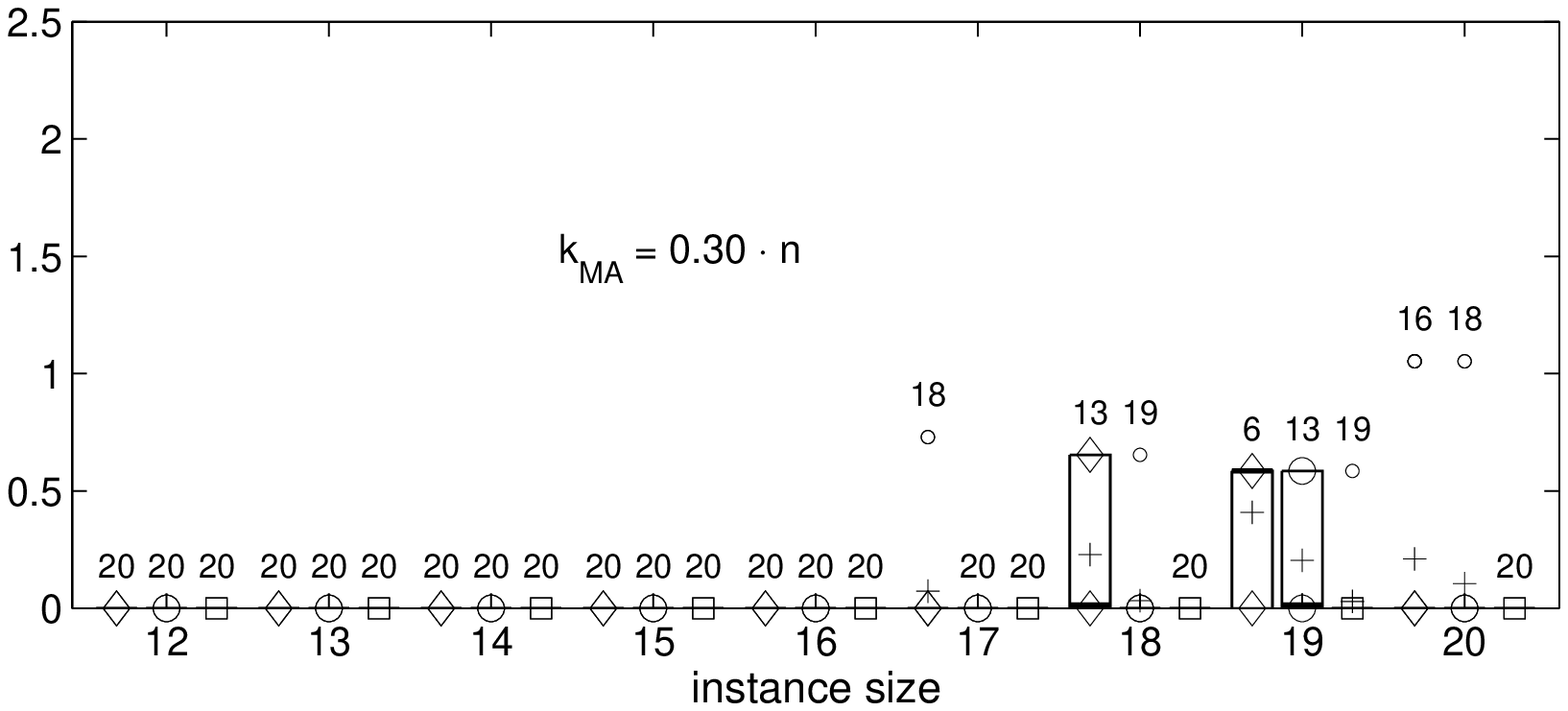,scale=.50}
\end{array}%
\renewcommand{\arraystretch}{0}%
$ \caption[Relative distances to optimum for different arities for
{\protect\linebreak[4]} \BSMABEMB~and $K_{MA} \in \{0.3 \cdot n, 0.5
\cdot n, 0.75 \cdot n\}$, for sizes ranging from 12 up to
20.]{Relative distances to optimum for different arities for
\BSMABEMB~and $K_{MA} \in \{0.3 \cdot n, 0.5 \cdot n, 0.75 \cdot
n\}$, for sizes ranging from 12 up to 20. Each box summarizes 20
runs.} \label{stilllife:fig:BS-MA-BE-MB-kMA}
\end{figure}

\begin{figure}[h!]
\centering
\epsfig{file=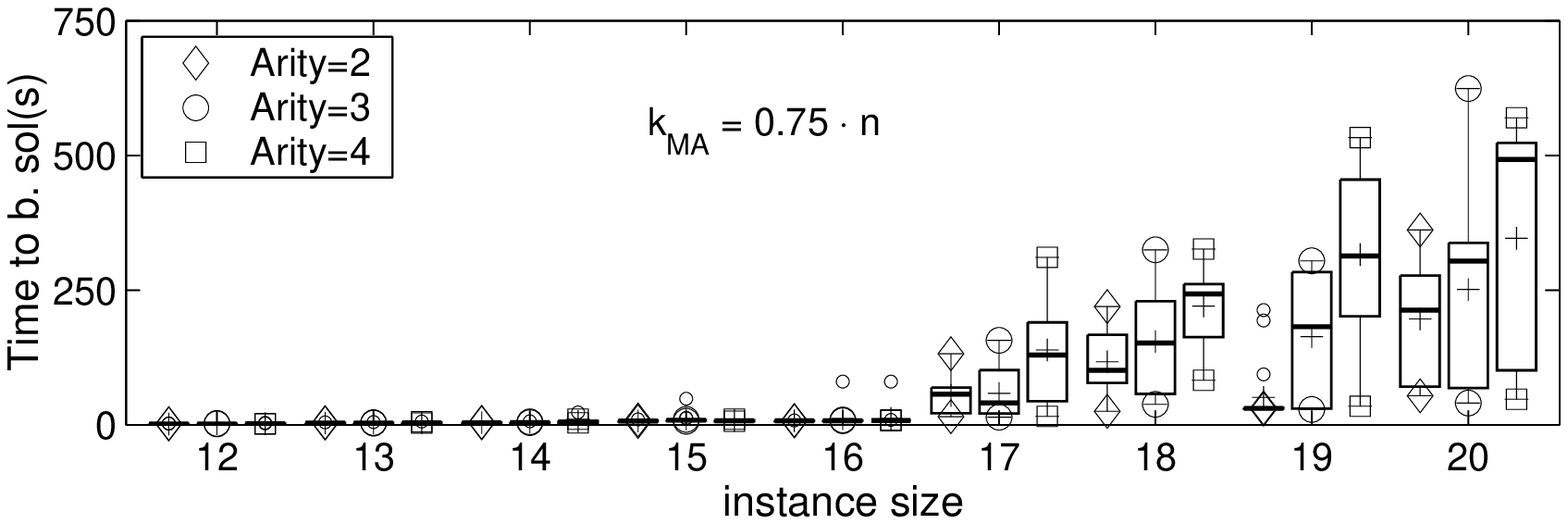,scale=.50}
\caption[Time  to best solution for different arities for
\BSMABEMB~and $k_{MA}=0.75 \cdot n$, for sizes ranging from 12 up to
20.]{Time (in seconds) to best solution for different arities for
\BSMABEMB~and $k_{MA}=0.75 \cdot n$, for sizes ranging from 12 up to
20. Each box summarizes 20 runs.}
\label{stilllife:fig:BS-MA-BE-MB-75-TIME}
\end{figure}

Experiments were repeated for the hybrid algorithm equipped with the
new lower bound, \BSMABEMB. Fig.~\ref{stilllife:fig:BS-MA-BE-MB-kMA}
shows the results of these experiments for values of $k_{MA} \in
\{0.3 \cdot n, 0.5 \cdot n, 0.75 \cdot n\}$. The algorithm finds the
optimum for all instances and arities and the relative distance to
the optimum for the worst solution found is less than 1.05\% in all
cases. The best results are obtained with $arity=4$, although this
requires slightly more execution time. Note also how \BSMABEMB~is
less sensitive to the setting of parameter $k_{MA}$, which means
that execution times can be reduced considerably using a large value
for this parameter (see
Fig.~\ref{stilllife:fig:BS-MA-BE-MB-75-TIME}). The particular
combination of parameters $k_{MA}=0.75 \cdot n$ and $arity=4$
provides excellent results at a lower computational cost, as
execution times are always below 570 seconds for $n \leqslant 20$.
As a comparison, recall that the only approach in the literature
that can solve these instances -- described by \citeA{Larrosa05} --
requires over 33 minutes for $n=18$, 15 hours for $n=19$ and 2 days
for $n=20$, and that other approaches are unaffordable for $n>15$.
Note however that these times correspond to a computational platform
different to ours. In order to do a fairer comparison, we executed
the algorithm of Larrosa et al.~\footnote{Available at
\url{http://www.lsi.upc.edu/~larrosa/publications/LIFE-SOURCE-CODE.tar.gz}
. Time for $n=19$ could not be obtained as the code provided by
Larrosa et al. can only be used with even sized instances.} in our
platform. In this case, it required 1867 seconds (i.e., more than 31
minutes) in order to solve the $n=18$ instance, and more than 1 day
and 18 hours to solve the $n=20$ instance. These values are very
close to the times reported by \citeA{Larrosa05}, and hence indicate
that the computational platforms are fairly comparable.

\begin{figure}[h!]
\centering
\epsfig{file=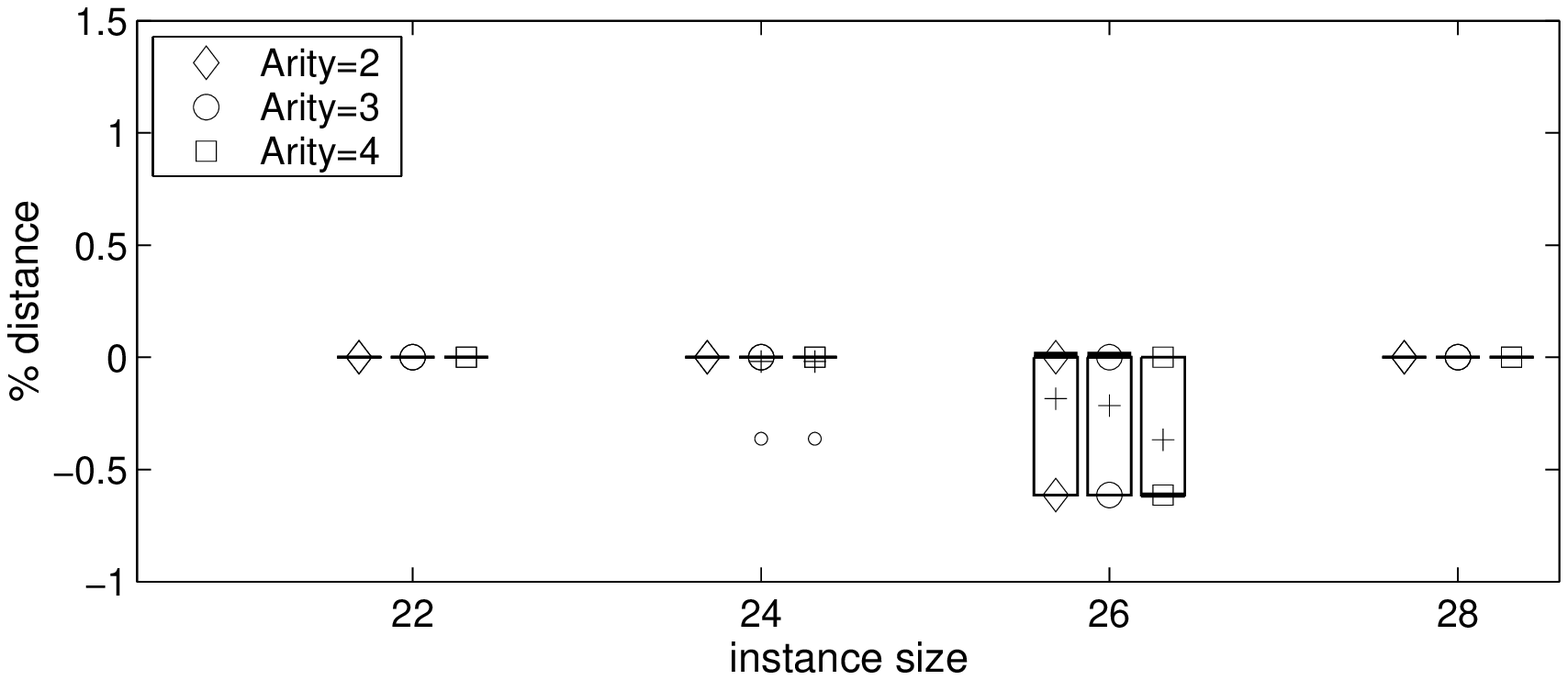,scale=.50}
\caption[Relative distances to best known solutions for different
arities for \BSMABEMB~and $k_{MA}=0.3 \cdot n$, for very large
instances.]{Relative distances to best known solutions for different
arities for \BSMABEMB~and $k_{MA}=0.3 \cdot n$, for very large
instances (i.e., sizes of 22, 24, 26, and 28). Each box summarizes
20 runs. Note the improvement of best known solutions for sizes 24
and 26.} \label{stilllife:fig:BS-MA-BE-MB-21-28}
\end{figure}

\begin{figure}[h!]
    \centering
    \epsfig{file=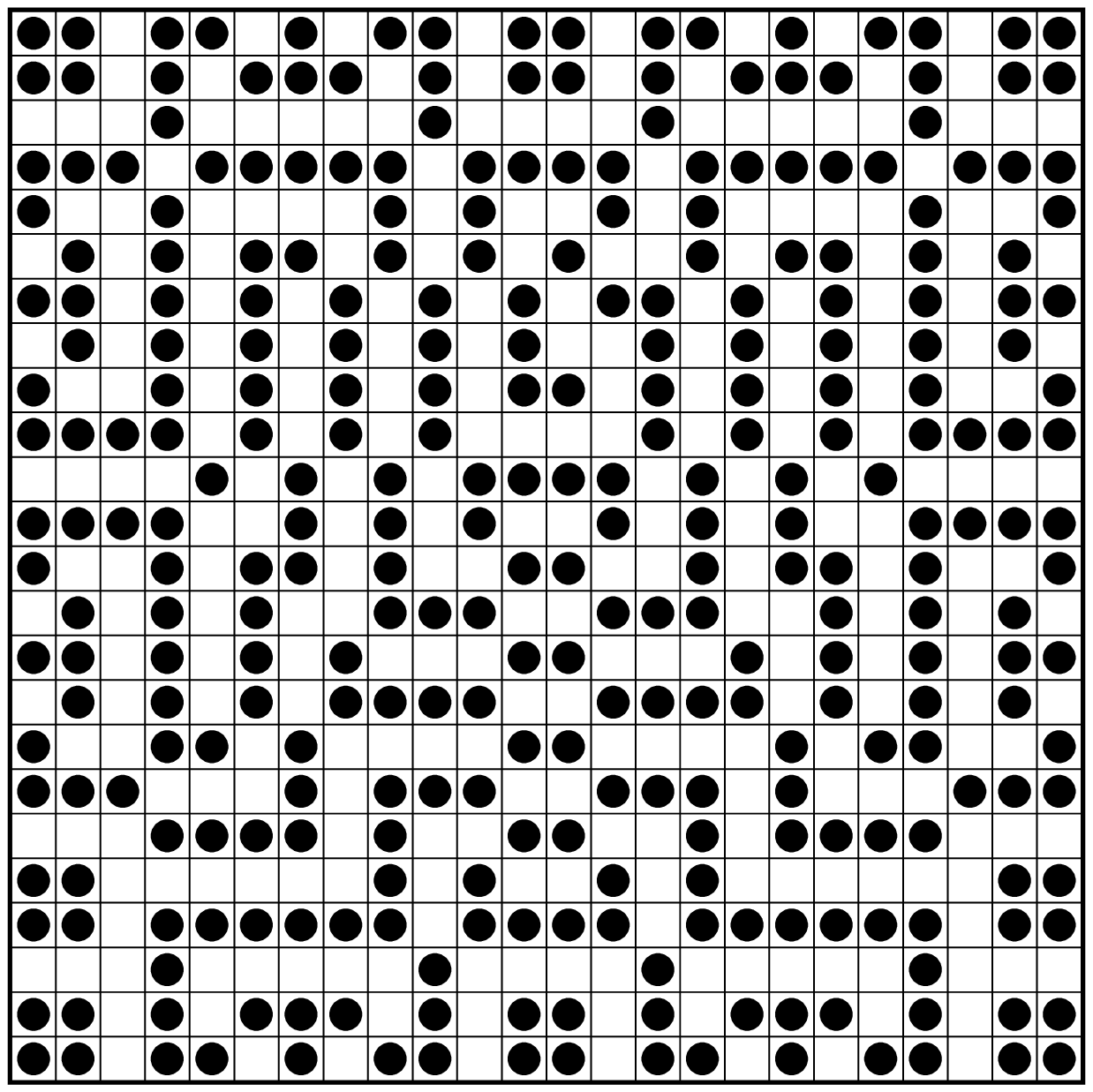,scale=.5}
    \hspace{3mm}
    \epsfig{file=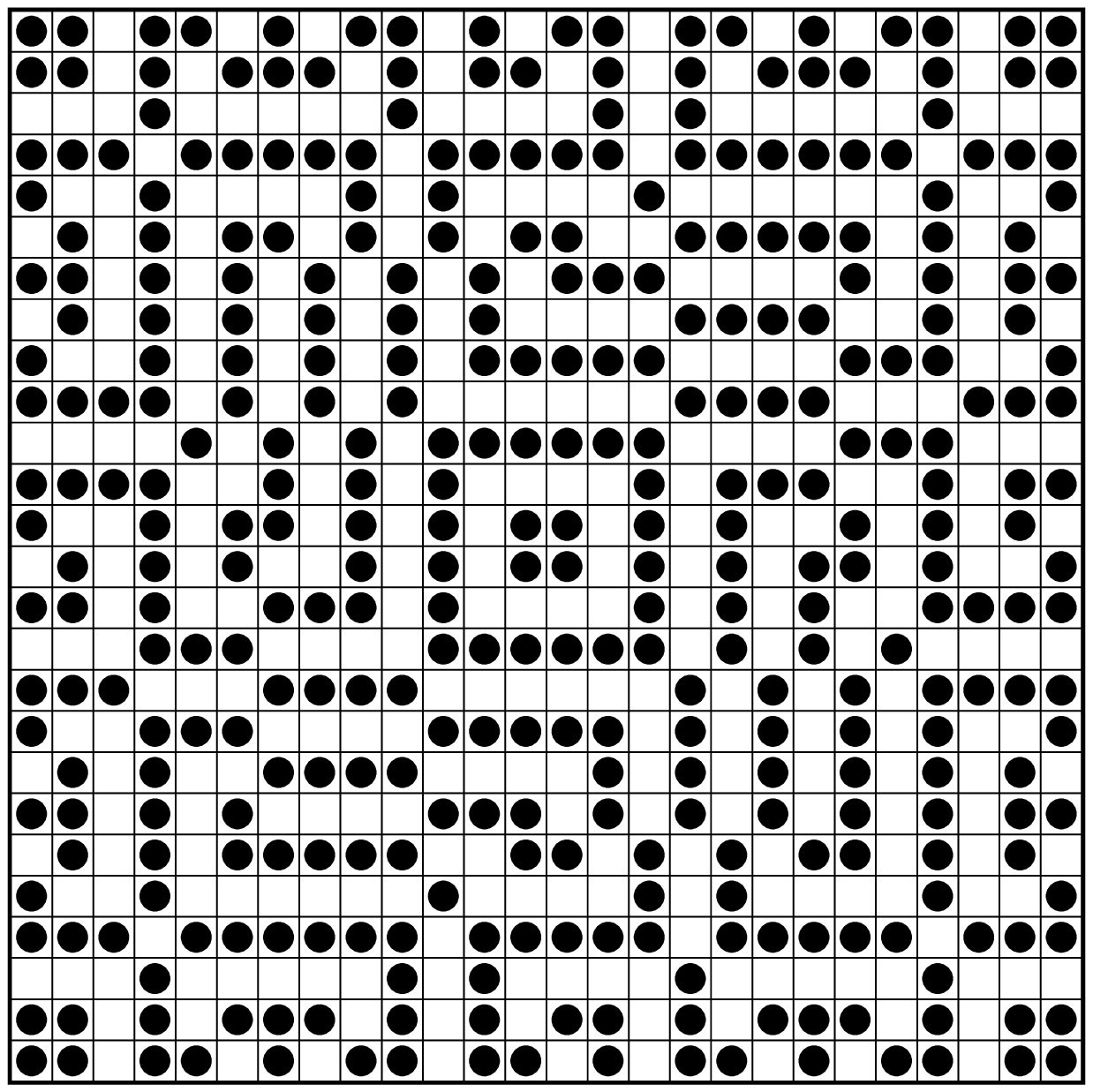,scale=.54}
    \caption{New best known maximum density still lifes for $n \in \{24,26\}$.}
    \label{stilllife:fig:new:sols}
\end{figure}

\begin{table}[h!]
\caption{Optimal solutions for the SMDLP.}
\label{stilllife:table:SMDSLP} \centering
\begin{tabular} {*{10}{r@{\ \ \ }}}
\\
\hline
$n$ \vline\hfill  & 12 & 13 & 14 &  15 &  16 &  17 &  18 &  19 &  20\\
\hline
opt \vline\hfill & 68 & 79 & 92 & 106 & 120 & 137 & 154 & 172 & 192\\
\hline\\
\\
\cline{1-5}
$n$ \vline\hfill  & 22 & 24  & 26  & 28  \\
\cline{1-5}
opt \vline\hfill & 232 & 276 & 326 & 378 \\
\cline{1-5}
\end{tabular}
\end{table}

\subsection{Results on Very Large Instances}
As already mentioned, there is currently no approach available to
tackle the \MDSLP\ for $n>20$. \citeA{Larrosa05} tried their
algorithm for $n=21$ and $n=22$, but they could not solve any of
those instances within a week of CPU). For these very large
instances, only solutions to some relaxations of the problem are
known. One of these relaxations, known as the symmetrical maximum
density still life problem (SMDSLP), was proposed in
\cite{bosch:three-life-designs-cpaior02}, and consists of
considering only symmetric boards (either horizontally or
vertically) which reduces the search space from $2^{n^2}$ to
$2^{n\lceil n/2 \rceil}$.

The optimized version of BE algorithm (Section
\ref{stilllife:sect:BE}) can be used find vertically symmetric still
lifes, by defining as the variable domain, a set that contains only
symmetric values for rows,
\begin{equation}
{\cal D} = \{r\ {\bf or}\ \overline{r}\ |\ r \in \{0 \dotdot
2^{\lceil n/2 \rceil}-1\}\}.
\end{equation}
\citeA{DBLP:conf/cp/LarrosaM03} and \citeA{Larrosa05} used this
algorithm to solve the SMDSLP for the instances considered so far in
this paper (i.e., for $n \in \{12 \dotdot 20\}$), as well as for
very large instances (i.e., $n \in \{ 22,24,26,28\}$). Results are
summarized in Table~\ref{stilllife:table:SMDSLP}, which shows for
each instance size the optimal symmetrical solution (as the number
of dead cells). Clearly, the costs of optimal symmetric still lifes
are upper bounds for the \MDSLP, that can additionally be observed
to be very tight for $n\leqslant 20$. Results for $n>20$ are
currently the best known solutions for these instances.

We also run our algorithm (\BSMABEMB) for these very large instances
(i.e., $n \in \{22,24,26, 28\}$), and compare our results to
symmetrical solutions for  these instances. Results (displayed in
Fig.~\ref{stilllife:fig:BS-MA-BE-MB-21-28}  shows that our algorithm
was able to find two new best known solutions for the \MDSLP, namely
for $n=24$ and $n=26$. There are 275 and 324 dead cells respectively
in the new solutions. These solutions are pictured in
Fig.~\ref{stilllife:fig:new:sols}. Incidentally, our algorithm could
also find a solution with 325 dead cells for the $n=26$ instance.
For the other instances, our algorithm could reach the best known
solutions consistently. Let us note that the computation of
mini-Buckets for these very large instances was done by considering
four clustered costs functions for variables in each row of the
board, as the complexity when using three costs functions was still
too high.

\section{Conclusions}
\label{stilllife:sect:conclusions} The \MDSLP\ represents an
excellent example of WCSP; its highly constrained nature is typical
in many optimization scenarios. Furthermore, the algorithmic
hardness of solving this problem illustrates the limitations of
classical optimization approaches. For this reason, it is not
surprising that this problem has attracted the interest of the
constraint-programming community, and has been central in the
development and assessment of sophisticated techniques such as
bucket elimination (\BE). However, the high space complexity of BE
as an exact technique \cite{dechter:bucket-elimination-AI99}, makes
this approach impractical for large instances. In this work, we have
presented several proposals for the hybridization of \BE\ with
memetic algorithms and beam search (BS), and showed that they
represent very promising models. The experimental results have been
very positive, solving to optimality large instances of the \MDSLP.
We have also studied the influence that  multi-parent recombination
have on the performance of the algorithm. The results indicate
multi-parent recombination can help to improve the results obtained
by previous approaches.

Among all our proposals, we must distinguish a new algorithm
resulting from the hybridization, at different levels, of complete
solving techniques (i.e., bucket elimination), incomplete
deterministic methods (i.e., beam search and mini-buckets) and
stochastic algorithms (i.e., memetic algorithms), that empirically
produces good-quality results, not only solving to optimality very
large instances of the constrained problem in a relatively short
time, but also providing new best known solutions in some large
instances. This algorithm exploits the technique of the mini-buckets
to compute tight yet computationally inexpensive lower bounds of the
partial solutions that are considered in the BS part of the hybrid
algorithm.

As future work, we plan to consider complete versions of the
hybrid algorithm. This involves the use of adequate data
structures to store not yet considered but promising
branch-and-bound nodes. While the memory requirements will of
course grow enormously with the size of the problem instance
considered, it will be interesting to analyze the computational
tradeoffs of the algorithm as an anytime technique.

\section*{Acknowledgments}
We would like to thank Javier Larrosa for his valuable comments,
that helped us to improve significantly a preliminary version of
this paper, and for permitting us to utilize his notation in the
preliminaries of this work. We would also like to acknowledge the
support of the Spanish MCyT under grant TIN2005-08818-C04-01.

\vskip 0.2in
\bibliography{arxiv}
\bibliographystyle{theapa}

\end{document}